\documentclass[10pt,journal,compsoc]{IEEEtran}

\usepackage[USenglish]{babel}
\usepackage{graphicx} % necessary for figures in pdf_tex format
\graphicspath{{./images/}{./plots/}} % tell the package where too look; necessary when using subfolders and pdf_tex files
\usepackage[usenames,svgnames,table,rgb,dvipsnames]{xcolor} % necessary for figures in pdf_tex format
\usepackage[]{mathtools} % loads amsmath; former options: nosumlimits,nonamelimits
\mathtoolsset{showonlyrefs} % provided by mathtools; can place the number differently

\usepackage{subcaption} % for subfigures
\usepackage[hang]{footmisc}
\usepackage[export]{adjustbox} % align included images vertically at the top 
\usepackage{enumitem} % for nice list options
\usepackage{wrapfig}
\usepackage{tabularx}
\usepackage{colortbl}
\usepackage{booktabs}
\usepackage{multirow}
\usepackage{leftidx} % usage: \leftidx{<left>}{<base>}{<right>}
\usepackage{csquotes}
\usepackage{pgf}

\usepackage[lined,linesnumbered,ruled]{algorithm2e}

\iffalse
	\usepackage[
	style=ieee,
	backend=bibtex,%bibtex | biber
	backref=false,%set to true for "(on page X)"
	natbib=true,
	maxnames=4,
	minnames=1,
	url=false, 
	doi=false,
	eprint=false
	]{biblatex}
\else
	\usepackage[numbers]{natbib}
\fi

\usepackage[nolist,nohyperlinks]{acronym}
\acrodef{ADaPT}{Adaptive Policy Transfer for Stochastic Dynamics}
\acrodef{ADN}{Activation Dynamics Network}
\acrodef{ARC-t}{Asymmetric Regularized Cross-domain transformation}
\acrodef{A2RP}{Averaged Two-Replication Procedure}
\acrodef{ARPL}{Adversarially Robust Policy Learning}
\acrodef{BO}{Bayesian Optimization}
\acrodef{BDR}{Bayesian Domain Randomization}
\acrodef{BDA}{Bayesian Domain Adaptation}
\acrodef{CMA}{Covariance Matrix Adaptation}
\acrodef{CMA-ES}{Covariance Matrix Adaptation - Evolutionary Strategies}
\acrodef{CNN}{Convolutional Neural Network}
\acrodef{CoM}{Center of Mass}
\acrodef{CVaR}{Conditional Value at Risk}
\acrodef{DA}{Domain Adaptation}
\acrodef{DAN}{Deep Adaptation Network}
\acrodef{DDPG}{Deep Deterministic Policy Gradient}
\acrodef{DMP}{Dynamic Movement Primitive}
\acrodefplural{DMP}[DMPs]{Dynamic Movement Primitives}
\acrodef{DNN}{Deep Neural Network}
\acrodef{DoF}{Degree of Freedom}
\acrodefplural{DOF}[DOF]{Degrees of Freedom}
\acrodef{DR}{Domain Randomization}
\acrodef{DS}{Dynamical System}
\acrodef{EoM}{Equations of Motion}
\acrodef{EPOpt}{Ensemble Policy Optimization}
\acrodef{FNN}{Feedforward Neural Network}
\acrodef{FPO}{Fingerprint Policy Optimization}
\acrodef{GAN}{Generative Adversarial Network}
\acrodef{GMM}{Gaussian Mixture Model}
\acrodef{GP}{Gaussian Process}
\acrodef{GPS}{Guided Policy Search}
\acrodef{HER}{Hindsight Experience Replay}
\acrodef{HRI}{Honda Research Institute}
\acrodef{HRIE}{Honda Research Institute Europe}
\acrodef{I2RP}{Independent Two-Replication Procedure}
\acrodef{LTI}{Linear Time-Invariant}
\acrodef{LQR}{Linear-Quadratic Regulator}
\acrodef{LSDA}{Large Scale Detection through Adaptation}
\acrodef{LSTM}{Long Short-Term Memory}
\acrodef{LWPR}{Locally Weighted Projection Regression}
\acrodef{MDP}{Markov Decision Process}
\acrodef{MP}{Movement Primitive}
\acrodefplural{MDP}{Markov Decision Processes}
\acrodef{MRP}{Multiple Replications Procedure}
\acrodef{MuJoCo}{Multi-Joint dynamics with Contact}
\acrodef{NN}{Neural Network}
\acrodef{NPG}{Natural Policy Gradient}
\acrodef{ODE}{Ordinary Differential Equation}
\acrodef{ODEphys}[ODE]{Open Dynamics Engine}
\acrodef{OG}{Optimality Gap}
\acrodef{PNN}{Progressive Neural Network}
\acrodef{PD}{Proportional-Derivative}
\acrodef{POMDP}{Partially-Observable Markov Decision Process}
\acrodef{PPO}{Proximal Policy Optimization}
\acrodef{ProMP}{Probabilistic Movement Primitive}
\acrodef{QBB}{Quanser Ball-Balancer}
\acrodef{QCP}{Quanser Cart-Pole}
\acrodef{QQ}{Quanser Qube}
\acrodef{RA}{Retrospective Approximation}
\acrodef{RARL}{Robust Adversarial Reinforcement Learning}
\acrodef{REPS}{Relative Entropy Policy Search}
\acrodef{RGB}{Red Green Blue}
\acrodef{RNN}{Recurrent Neural Network}
\acrodef{RL}{Reinforcement Learning}
\acrodef{SAA}{Sample Average Approximation}
\acrodef{SOB}{Simulation Optimization Bias}
\acrodef{SP}{Stochastic Program}
\acrodefplural{SP}{Stochastic Programs}
\acrodef{SPOTA}{Simulation-based Policy Optimization with Transferability Assessment}
\acrodef{TRPO}{Trust Region Policy Optimization}
\acrodef{UCB}{Upper Confidence Bound}
\acrodef{UCBOG}{Upper Confidence Bound on the Optimality Gap}
\acrodef{UCSOB}{Upper Confidence bound on the Simulation Optimization Bias}
\acrodef{UP-OSI}{Universal Policy - Online System Identification}

\definecolor{fMRTblack}{HTML}{2B2E34}
\definecolor{fMRTwhite}{HTML}{F3EEE1}
\definecolor{fMRTyellow}{HTML}{FFD564}
\definecolor{fMRTorange}{HTML}{FF5500}
\colorlet{fMRTlightGray}{white!80!fMRTblack}
\colorlet{fMRTgray}{white!35!fMRTblack}
\colorlet{fMRTdarkGray}{white!20!fMRTblack}
\colorlet{fMRTlightOrange}{fMRTorange!60!fMRTwhite}
\definecolor{fMRTblue}{HTML}{3333B3}
\colorlet{fMRTlightBlue}{fMRTblue!30!white}
\definecolor{fMRTverylightRed}{HTML}{FF8E6B}
\colorlet{fMRTlightRed}{red!60!fMRTwhite}
\colorlet{fMRTdarkRed}{red!70!fMRTgray} % red!60!fMRTgray
\definecolor{fMRTlightGreen}{HTML}{8CDD81} % green soap	
\definecolor{fMRTborwn}{HTML}{8B4513} % saddle brown
\colorlet{fMRTlightBorwn}{fMRTborwn!60!fMRTwhite}

\colorlet{fMRTdark@backgroundInner}{fMRTblack}
\colorlet{fMRTdark@backgroundOuter}{fMRTblack}
\colorlet{fMRTlight@backgroundInner}{fMRTwhite}
\colorlet{fMRTlight@backgroundOuter}{fMRTwhite}

\newcolumntype{L}{>{\raggedright\arraybackslash}X}
\newcolumntype{R}{>{\raggedleft\arraybackslash}X}
\newcolumntype{C}{>{\centering\arraybackslash}X}

\usepackage{xifthen}
\newcommand{\RR}{\mathbb{R}} % the set of real numbers to the power of #1
\newcommand{\onehalf}{\frac{1}{2}} % 1/2
\newcommand{\onefourth}{\frac{1}{4}} % 1/4
\newcommand{\alphad}{\dot{\alpha}} % not bold
\newcommand{\stateset}[1][]{\mathcal{S}_{#1}} % set of states in MDP; Note: using \stateset as a subscript requires curly brackets, i.e. _{\stateset}
\newcommand{\actionset}[1][]{\mathcal{A}_{#1}} % set of actions in MDP
\newcommand{\transprob}[1][]{\mathcal{P}_{#1}} % transition function in MDP
\newcommand{\initstatedistr}[1][]{\mu_{0#1}} % initial state distribution in MDP
\newcommand{\domparamdistrsym}{\nu} % prob dist for the physics params
\newcommand{\domparamdistr}[1]{\domparamdistrsym\!\left( #1\right) }
\newcommand{\domdistrparam}{\phi} % parameter of the distribution of the domain params
\newcommand{\domdistrparams}{\fphi} % parameters of the distribution of the domain params
\newcommand{\mdp}[1][]{\mathcal{M}_{#1}} % initial state distribution in MDP
\newcommand{\dimstate}{{n_s}} % d
\newcommand{\dimact}{{n_a}} % p
\newcommand{\dimpolparam}{{n_{\polparam}}} % n
\newcommand{\dimdomparam}{{n_{\domparam}}} % m
\newcommand{\optsym}{\star} % symbol to indicate optimality

\newcommand{\candsym}{{n_c}}
\newcommand{\refsym}{{n_r}}
\usepackage{xspace}
\newcommand{\etal}{et al.~}
\newcommand{\eg}{e.g.\xspace}
\newcommand{\ie}{i.e.\xspace}

\newcommand{\iid}{i.i.d.~}

\newcommand{\wrt}{w.r.t.~}
\newcommand{\aka}{a.k.a.~}

\newcommand{\simtosim}{{sim-to-sim}\xspace}

\newcommand{\simtoreal}{{sim-to-real}\xspace}
\newcommand{\sota}{{state-of-the-art}\xspace}

\newcommand{\ballbalancersym}{bb}
\newcommand{\cartpolesym}{cp}

\usepackage{amsopn} % for \DeclareMathOperator
\DeclareMathOperator*{\argmax}{arg\,max}

\usepackage{amssymb} % for \mathbb
\newcommand{\distr}[3]{\mathcal{#1}\left( #2 \big| #3\right) } % probability distribution: #1 symbol, #2 random variable, 3# parameters
\newcommand{\distrnormal}[2]{\distr{\mathcal{N}}{#1}{#2}} % normal distribution: #1 random variable, #2 parameters, e.g. mean, covar
\newcommand{\distruniform}[2]{\distr{\mathcal{U}}{#1}{#2}} % uniform distribution: #1 random variable, #2 parameters, e.g. min, max
\newcommand{\distrbern}[2]{\distr{\mathcal{B}}{#1}{#2}} % Bernoulli distribution: #1 random variable, #2 parameters, e.g. p
\newcommand{\Esub}[2]{\mathbb{E}_{#1} \! \left[ #2\right] } % expected value with subscript
\newcommand{\Esubbig}[2]{\mathbb{E}_{#1} \! \big[ #2\big] } % expected value with subscript (enforce big brackets)
\newcommand{\EsubBig}[2]{\mathbb{E}_{#1} \! \Big[ #2\Big] } % expected value with subscript (enforce Big brackets)
\renewcommand{\Pr}[1]{\mathbb{P}\left( {#1}\right) } % probability of an event
\newcommand{\quantilesym}{Q} % or \mathrm{Q} or \mathbb{Q}
\newcommand{\quantile}[2]{\quantilesym_{#1} \! \left[ #2\right] } % quantile of a set: #1 level, #2 set
\newcommand{\bias}[1]{\mathrm{b}[#1] } % bias | former \left[#1 \right]
\renewcommand{\ln}[1]{\mathrm{ln}\left( #1\right) } % logarithm
\renewcommand{\log}[1]{\mathrm{log}\left( #1\right) } % logarithm
\renewcommand{\exp}[1]{\mathrm{exp}\left( #1\right) } % exponential function

\renewcommand{\sin}[1]{\mathrm{sin}\!\left( #1\right) }
\renewcommand{\cos}[1]{\mathrm{cos}\!\left( #1\right) }
\makeatletter
\newcommand{\@givennostar}[2]{\left. #1\right| #2} % #1 = random variable, #2 = condition
\newcommand{\@givenstar}[3][]{#2 #1| #3} % #1(optional) = size of dilimiter, #2 = random variable, #3 = condition, e.g. \given*{\small}{a}{b}
\newcommand{\given}{\@ifstar\@givenstar\@givennostar}
\makeatother
\usepackage{mathtools}
\DeclarePairedDelimiter{\ceil}{\lceil}{\rceil} % larger integer
\DeclarePairedDelimiter{\floor}{\lfloor}{\rfloor} % samller integer

\usepackage[%
detect-all,%
scientific-notation=false%
]{siunitx}
\newcommand{\tran}{^\textsf{\upshape T}} % transposed
\usepackage{xparse}

\newcommand{\rewsym}{r}
\DeclareDocumentCommand{\rewfcn}{O{} O{} m}{\rewsym_{#1}^{#2}\!\left( #3\right) }

\newcommand{\trajssym}{\ftau}
\DeclareDocumentCommand{\traj}{O{} O{}}{\trajssym_{#1}^{#2}} % 2 optional args
\newcommand{\statedistrsym}{\mu}
\DeclareDocumentCommand{\statedistr}{O{} O{} m}{\statedistrsym_{#1}^{#2}\!\left( #3\right) } % steady state distribution

\newcommand{\domparam}{\xi}
\newcommand{\domparams}{\fxi}
\newcommand{\domparamsnom}{\bar{\fxi}}

\newcommand{\polsym}{\pi}
\DeclareDocumentCommand{\pol}{O{} O{} m}{\polsym_{#1}^{#2}\!\left( #3\right) } % policy; optional arg for dependency on physics params
\newcommand{\polparam}{\theta}
\newcommand{\polparamspace}{\Theta}
\DeclareDocumentCommand{\polparams}{O{} O{}}{\ftheta_{#1}^{#2}} % 2 optional args; NOTE: max_\polparams yiels an error, but max_{\polparams} is fine
\newcommand{\solcandi}{\polparams[][c]} % solution candidate
\DeclareDocumentCommand{\polparamsopt}{O{} O{}}{\ftheta_{#1}^{#2 \optsym}} % optimal policy parameters indicated by star superscript
\DeclareDocumentCommand{\polparamopt}{O{} O{}}{\polparam_{#1}^{#2 \optsym}} % optional arg is the index (not perfect - it would be better to define a new \DeclareDocumentCommand)
\DeclareDocumentCommand{\polparamsopt}{O{} O{}}{\polparams[#1][#2 \optsym]} % optional arg is the index (not perfect - it would be better to define a new \DeclareDocumentCommand)
\newcommand{\polparamsoptcand}{\polparams[\candsym][\optsym]} % opt cand sol
\newcommand{\optgapsym}{G}
\DeclareDocumentCommand{\optgap}{O{} O{} m}{\optgapsym_{#1}^{#2}\!\left( #3\right) } % optimality gap of the policy #3; e.g. \optgap[1][2]{3}
\newcommand{\estoptgapsym}{\hat{G}}
\DeclareDocumentCommand{\estoptgap}{O{} O{} m}{\estoptgapsym_{#1}^{#2}\!\left( #3\right) } %  estimator of the optimality gap of the policy #3 calculated from #1 samples; e.g. \optgap[1][2]{3}
\newcommand{\meanoptgapsym}{\bar{G}}
\DeclareDocumentCommand{\optgapmean}{O{} O{} m}{\meanoptgapsym_{#1}^{#2}\!\left( #3\right) } % sample mean of the optimality gap from n_G calucations of G_n
\newcommand{\optgapsampsetsym}{\mathcal{G}}
\DeclareDocumentCommand{\optgapsampset}{O{} O{}}{\optgapsampsetsym_{#1}^{#2}} % set of optimality gap samples

\newcommand{\sob}[1]{\mathrm{b}\!\left[#1 \right]}

\newcommand{\sobNormal}[1]{\mathrm{b} [#1]}

\newcommand{\numbssamples}{{n_b}}
\iftrue
\newcommand{\bootsym}{B}
\DeclareDocumentCommand{\bootestoptgap}{O{} O{} m}{\estoptgapsym_{#1}^{\bootsym #2}\!\left( #3\right) }
\DeclareDocumentCommand{\bootmeanoptgap}{O{} O{} m}{\meanoptgapsym_{#1}^{\bootsym #2}\!\left( #3\right) }
\DeclareDocumentCommand{\bootoptgapsampset}{O{} O{}}{\optgapsampsetsym_{#1}^{\bootsym #2}}
\fi

\newcommand{\edrsym}{J}
\DeclareDocumentCommand{\edr}{O{} m}{\edrsym_{#1}\!\left( #2\right)} % expected discounted return

\newcommand{\estedrsym}{\hat{J}}
\DeclareDocumentCommand{\estedr}{O{} m}{\estedrsym_{#1}\!\left( #2\right)} % estimator of the expected (trajs) discounted return
\newcommand{\eedrsym}{J} % former \bar{J}
\DeclareDocumentCommand{\eedr}{O{} m}{\eedrsym_{#1}\!\left( #2\right)} % expected (domains) expected (trajs) discounted return
\DeclareDocumentCommand{\esteedr}{O{} m}{\estedrsym_{#1}\!\left( #2\right)} % estimator of the expected (domains) expected (trajs) discounted return
\newcommand{\objfunsym}{f}
\DeclareDocumentCommand{\objfun}{O{} m}{\objfunsym_{#1}\!\left( #2\right)} % objective function

\newcommand{\dsdynsym}{f}
\DeclareDocumentCommand{\dsdyn}{O{} m}{\dsdynsym_{#1}\!\left( #2\right)} % 1 optional arg (subscript)

\newcommand{\potdynsym}{f}
\newcommand{\potsdynsym}{\ff}
\DeclareDocumentCommand{\potdyn}{O{} m}{\potdynsym_{#1}\!\left( #2\right)} % 1 optional arg (subscript)
\DeclareDocumentCommand{\potsdyn}{O{} m}{\potsdynsym_{#1}\!\left( #2\right)} % 1 optional arg (subscript)

\ExplSyntaxOn
\NewDocumentCommand{\eqsref}{m}{\quinn_eqsref:n {#1}}
\seq_new:N \l_quinn_eqsref_seq
\cs_new:Npn \quinn_eqsref:n #1
{
	\seq_set_split:Nnn \l_quinn_eqsref_seq { , } { #1 }
	\seq_pop_right:NN \l_quinn_eqsref_seq \l_tmpa_tl
	( % print the left parenthesis
	\seq_map_inline:Nn \l_quinn_eqsref_seq
	{ \ref{##1},\nobreakspace } % print the first references
	\exp_args:NV \ref \l_tmpa_tl % print the last or only one
	) % print the right parenthesis
}
\ExplSyntaxOff

\usepackage[lined]{algorithm2e}
\SetKw{kwNot}{not}
\SetKw{kwAnd}{and}
\SetKw{kwBreak}{break}
\SetKwInOut{Input}{input}
\SetKwInOut{Output}{output}
\SetKwInOut{Initialization}{init}
\SetKwComment{Comment}{$\triangleright$\ }{}
\SetKwRepeat{Do}{do}{while} % do ... while ... loop
\SetKwBlock{With}{with}{end} % usage: \With(condition){code}; see p.12 in https://mlg.ulb.ac.be/files/algorithm2e.pdf
\newcommand{\llIf}[2]{{\let\par\relax\lIf{#1}{#2}}}
\newcommand{\llElse}[1]{{\let\par\relax\lElse{#1}}}
\newcommand{\llFor}[2]{{\let\par\relax\lFor{#1}{#2}}}

\usepackage{bm}
\DeclareBoldMathCommand{\fnull}{0}
\DeclareBoldMathCommand{\fO}{0}
\DeclareBoldMathCommand{\fA}{A}
\DeclareBoldMathCommand{\fa}{a}
\DeclareBoldMathCommand{\fB}{B}
\DeclareBoldMathCommand{\fb}{b}
\DeclareBoldMathCommand{\fC}{C}
\DeclareBoldMathCommand{\fc}{c}
\DeclareBoldMathCommand{\fD}{D}
\DeclareBoldMathCommand{\fd}{d}
\DeclareBoldMathCommand{\fE}{E}
\DeclareBoldMathCommand{\fe}{e}
\DeclareBoldMathCommand{\fF}{F}
\DeclareBoldMathCommand{\ff}{f}
\DeclareBoldMathCommand{\fG}{G}
\DeclareBoldMathCommand{\fg}{g}
\DeclareBoldMathCommand{\fH}{H}
\DeclareBoldMathCommand{\fh}{h}
\DeclareBoldMathCommand{\fI}{I}
\DeclareBoldMathCommand{\fJ}{J}
\DeclareBoldMathCommand{\fK}{K}
\DeclareBoldMathCommand{\fM}{M}
\DeclareBoldMathCommand{\fm}{m}
\DeclareBoldMathCommand{\fN}{N}
\DeclareBoldMathCommand{\fn}{n}
\DeclareBoldMathCommand{\fo}{o}
\DeclareBoldMathCommand{\fP}{P}
\DeclareBoldMathCommand{\fp}{p}
\DeclareBoldMathCommand{\fQ}{Q}
\DeclareBoldMathCommand{\fq}{q}
\DeclareBoldMathCommand{\fq}{q}

\DeclareBoldMathCommand{\fR}{R}
\DeclareBoldMathCommand{\fr}{r}

\DeclareBoldMathCommand{\fS}{S}
\DeclareBoldMathCommand{\fs}{s}

\DeclareBoldMathCommand{\ft}{t}
\DeclareBoldMathCommand{\fT}{T}
\DeclareBoldMathCommand{\fU}{U}
\DeclareBoldMathCommand{\fu}{u}
\DeclareBoldMathCommand{\fV}{V}
\DeclareBoldMathCommand{\fv}{v}
\DeclareBoldMathCommand{\fW}{W}
\DeclareBoldMathCommand{\fw}{w}
\DeclareBoldMathCommand{\fx}{x}

\DeclareBoldMathCommand{\fY}{Y}
\DeclareBoldMathCommand{\fy}{y}
\DeclareBoldMathCommand{\fZ}{Z}
\DeclareBoldMathCommand{\falpha}{\alpha}

\DeclareBoldMathCommand{\fchi}{\chi}
\DeclareBoldMathCommand{\fepsilon}{\epsilon}
\DeclareBoldMathCommand{\fvarepsilon}{\varepsilon}
\DeclareBoldMathCommand{\fgamma}{\gamma}
\DeclareBoldMathCommand{\fGamma}{\Gamma}
\DeclareBoldMathCommand{\flambda}{\lambda}
\DeclareBoldMathCommand{\fLambda}{\Lambda}
\DeclareBoldMathCommand{\fmu}{\mu}
\DeclareBoldMathCommand{\fnu}{\nu}
\DeclareBoldMathCommand{\fomega}{\omega}
\DeclareBoldMathCommand{\fpi}{\pi}
\DeclareBoldMathCommand{\fphi}{\phi}
\DeclareBoldMathCommand{\fPhi}{\Phi}
\DeclareBoldMathCommand{\fpsi}{\psi}
\DeclareBoldMathCommand{\fsigma}{\sigma}
\DeclareBoldMathCommand{\fSigma}{\Sigma}
\DeclareBoldMathCommand{\ftau}{\tau}
\DeclareBoldMathCommand{\ftheta}{\theta}
\DeclareBoldMathCommand{\fTheta}{\Theta}

\DeclareBoldMathCommand{\fxi}{\xi}

\usepackage{url}
\title{Assessing Transferability from Simulation to Reality for Reinforcement Learning}
\date{}
\author{%
	Fabio~Muratore, %
	Michael~Gienger,~\IEEEmembership{Member,~IEEE}, %
	and~Jan~Peters,~\IEEEmembership{Fellow,~IEEE}%
	\IEEEcompsocitemizethanks{\IEEEcompsocthanksitem Fabio~Muratore and Jan~Peters are with the Intelligent Autonomous Systems Group, Technische Universit\"at Darmstadt, Germany.\protect\\
		Correspondence to fabio@robot-learning.de
	\IEEEcompsocthanksitem Fabio~Muratore and Michael~Gienger are with the Honda Research Institute Europe, Offenbach am Main, Germany.%
	\IEEEcompsocthanksitem Jan~Peters is with the Max Planck Institute for Intelligent Systems, T\"ubingen, Germany.}%
	\thanks{Manuscript received 21 June 2019; revised 2 October 2019.}} %  revised August 26, 2015

\begin{document}

% %%%%%%%%%%%%%%%%%%%%%%%%%%%%%%%%%%%%%%%%%%%%%%%%%% %
\IEEEtitleabstractindextext{%
\begin{abstract}
% %%%%%%%%%%%%%%%%%%%%%%%%%%%%%%%%%%%%%%%%%%%%%%%%%% %
% 1. Stating the problem: address the problem, but not the solution. You can include some background (a sentence or two about other approaches; yours or others) 
Learning robot control policies from physics simulations is of great interest to the robotics community as it may render the learning process faster, cheaper, and safer by alleviating the need for expensive real-world experiments.
% 2. Say why it is interesting: You may want to discuss importance. Remember that it’s not obvious to everyone else how important this problem is.
However, the direct transfer of learned behavior from simulation to reality is a major challenge. Optimizing a policy on a slightly faulty simulator can easily lead to the maximization of the `\acl{SOB}'~(SOB). In this case, the optimizer exploits modeling errors of the simulator such that the resulting behavior can potentially damage the robot.
% 3. Say what your solution is / what it achieves: You may want to discuss your insight (What did you discover? How did you approach the problem differently?).
We tackle this challenge by applying domain randomization, \ie, randomizing the parameters of the physics simulations during learning.
We propose an algorithm called \acf{SPOTA} which uses an estimator of the \acs{SOB} to formulate a stopping criterion for training.
The introduced estimator quantifies the over-fitting to the set of domains experienced while training.
% 4. Say what follows from your solution: What are the implications of your answer? Is it going to change the world, be a significant "win", be a nice hack, or simply serve as a road sign indicating that this path is a waste of time (all of the previous results are useful). Are your results general, potentially generalizable, or specific to a particular case? You should clearly say what your contribution is (Spell it out, do not assume we will read the paper carefully).
Our experimental results on two different second order nonlinear systems show that the new simulation-based policy search algorithm is able to learn a control policy exclusively from a randomized simulator, which can be applied directly to real systems without any additional training. % on the latter.
\end{abstract}

}

\maketitle
\IEEEdisplaynontitleabstractindextext
\IEEEpeerreviewmaketitle

% %%%%%%%%%%%%%%%%%%%%%%%%%%%%%%%%%%%%%%%%%%%%%%%%%% %
%\IEEEraisesectionheading
\section{Introduction}
\label{sec_introduction}
% %%%%%%%%%%%%%%%%%%%%%%%%%%%%%%%%%%%%%%%%%%%%%%%%%% %
% Training in reality vs. simulation
\IEEEPARstart{E}{xploration-based} learning of control policies on physical systems is expensive in two ways. For one thing, real-world experiments are time-consuming and need to be executed by experts. Additionally, these experiments require expensive equipment which is subject to wear and tear.
In comparison, training in simulation provides the possibility to speed up the process and save resources. A major drawback of robot learning from simulations is that a simulation-based learning algorithm is free to exploit any infeasibility during training and will utilize the flawed physics model if it yields an improvement during simulation. This exploitation capability can lead to policies that damage the robot when later deployed in the real world.
% Reality gap
The described problem is exemplary of the difficulties that occur when transferring robot control policies from simulation to reality, which have been the subject of study for the last two decades under the term `reality gap'.
Early approaches in robotics suggest using minimal simulation models and adding artificial \iid noise to the system's sensors and actuators while training in simulation \cite{Jakobi_etal_1995}. The aim was to prevent the learner from focusing on small details, which would lead to policies with only marginal applicability.
% Optimality gap
This over-fitting can be described by the \ac{SOB}, which is similar to the bias of an estimator.
The \ac{SOB} is closely related to the \ac{OG}, which has been used by the optimization community since the 1990s \cite{Hobbs_Hepenstal_1989,Mak_etal_1999}, but has not been transferred to robotics or \ac{RL}, yet.%
\begin{figure}[t]%
	\centering
	\includegraphics[height=3.8cm]{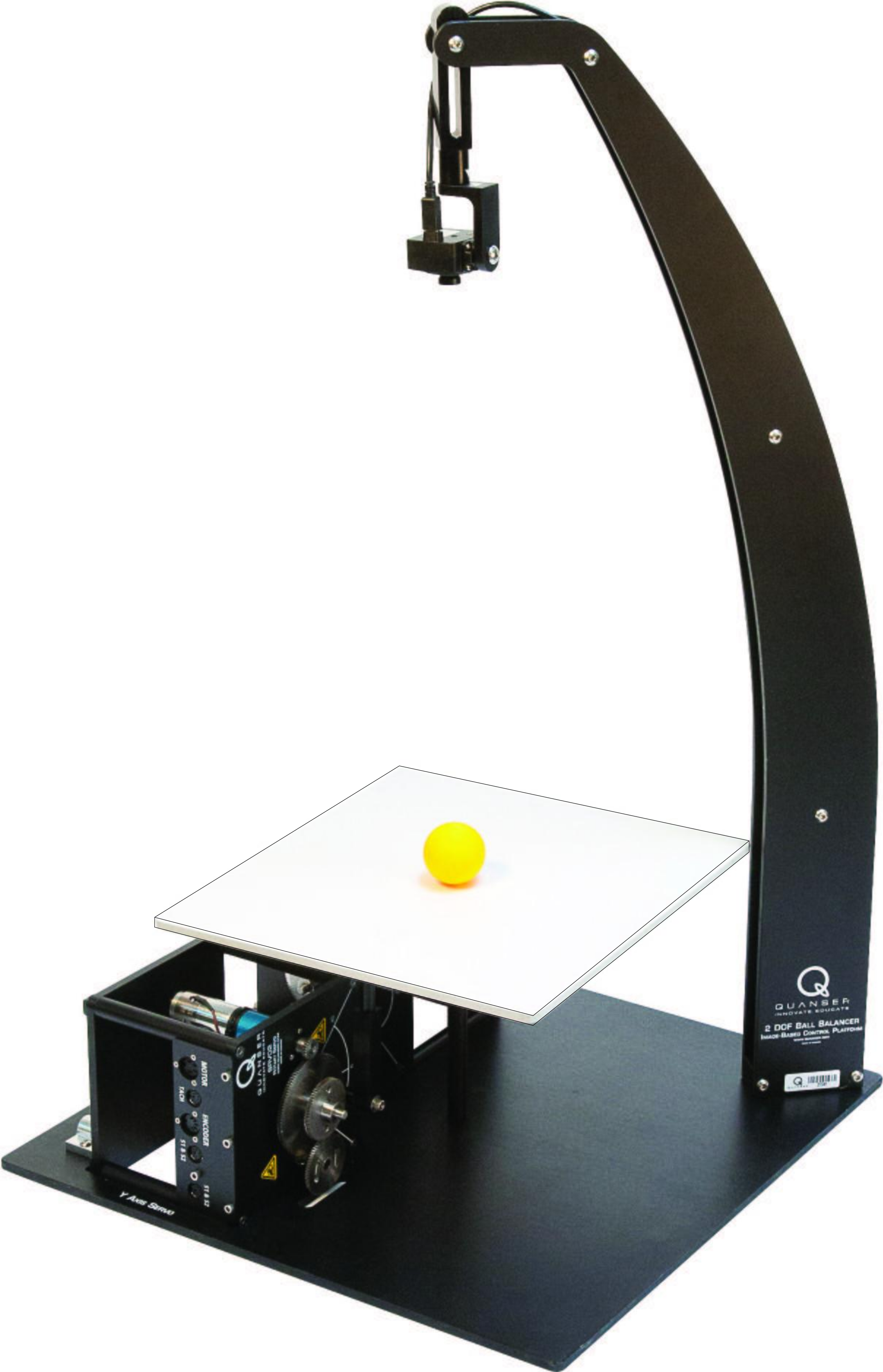}
	\hspace{1em}
	\includegraphics[height=3.3cm]{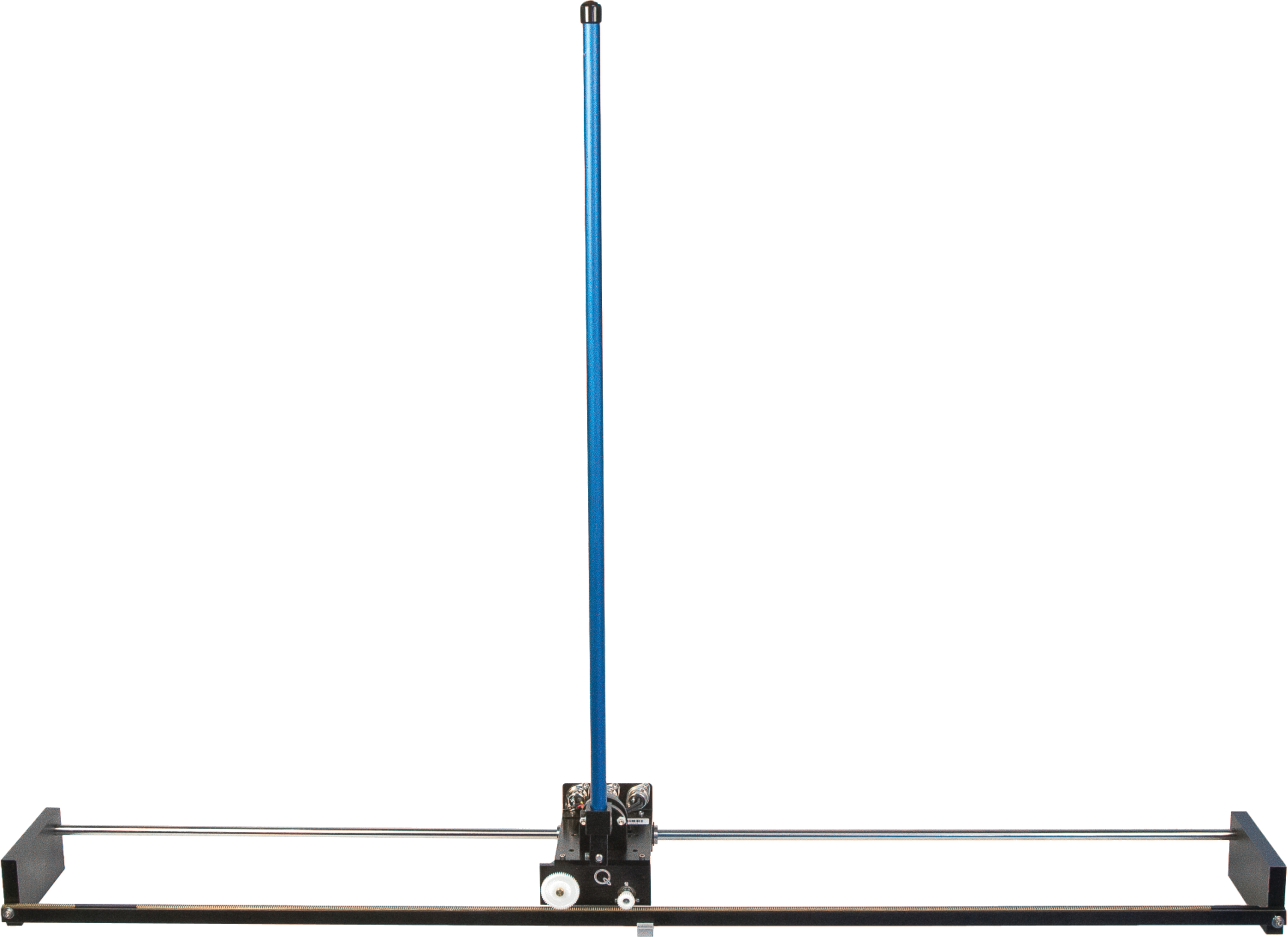}
	\caption{%
		Evaluation platforms by Quanser~\cite{Quanser_platforms}: (left) the \mbox{2 \acs{DoF}} Ball-Balancer, (right) the linear inverted pendulum, called Cart-Pole.
		Both systems are under-actuated nonlinear balancing problems with continuous state and action spaces.
		}
	\label{fig_Quanser_platforms}
\end{figure}%

% Growing interest in domain randomization
Deep \ac{RL} algorithms recently demonstrated super-human performance in playing games \cite{Mnih_etal_2015, Silver_etal_2017} and promising results in (simulated) robotic control tasks \cite{Lillicrap_etal_2015,Schulman_etal_2017,Rusu_etal_2017}. %James_Johns_2016
However, when transferred to real-world robotic systems, most of these methods become less attractive due to high sample complexity and a lack of explainability of state-of-the-art deep \ac{RL} algorithms.
As a consequence, the research field of domain randomization has recently been gaining interest \cite{Mordatch_etal_2015,Rajeswaran_etal_2017,Mandlekar_etal_2017,Pinto_etal_2017a,Pinto_etal_2017b,Peng_etal_2018,Yu_etal_2017,Chebotar_etal_2019}. This class of approaches promises to transfer control policies learned in simulation (source domain) to the real world (target domain) by randomizing the simulator's parameters (\eg, masses, extents, or friction coefficients) and hence train from a set of models instead of just one nominal model.
% Highlighting very recent successes
Further motivation to investigate domain randomization is given by the recent successes in robotic \simtoreal scenarios, such as the in-hand manipulation of a cube \cite{OpenAI_etal_2018}, swinging a peg in a hole, or opening a drawer \cite{Chebotar_etal_2019}.
% Bayesian point of view
The idea of randomizing the simulator's parameters is driven by the fact that the corresponding true parameters of the target domain are unknown. However, instead of relying on an accurate estimation of one fixed parameter set, we take a Bayesian point of view and assume that each parameter is drawn from an unknown underlying distribution.
Thereby, the expected effect is an increase in robustness of the learned policy when applied to a different domain. Throughout this paper, we use the term robustness to describe a policy's ability to maintain its performance under model uncertainties. In that sense, a robust control policy is more likely to overcome the reality gap.

% Bigger picture
Looking at the bigger picture, model-based control % (\eg, the linear-quadratic regulator)
only considers a system's nominal dynamics parameter values, while robust control %(\eg, $\Hinfty$ control) 
minimizes a system's sensitivity with respect to bounded model uncertainties, thus focuses the worst-case.
In contrast to these methods, domain randomization takes the whole range of parameter values into account. 

% Statement of the contributions
\textit{Contributions:} we advance the \sota by
\begin{enumerate}
	\item introducing a measure for the transferability of a solution, \ie, a control policy, from a set of source distributions to a different target domain from the same distribution,
	\item designing an algorithm which, based on this measure, is able to transfer control policies from simulation to reality without any real-world data, and
	\item validating the approach by conducting two \simtoreal experiments on under-actuated nonlinear systems.
\end{enumerate}
% Biref summary of this paper
The remainder of this paper is organized as follows:
we explain the necessary fundamentals (Section~\ref{sec_prob_statement}) for the proposed algorithm (Section~\ref{sec_SPOTA}). In particular, we derive the \acf{SOB} and the \acf{OG}.
After validating the proposed method in simulation, we evaluate it experimentally (Section~\ref{sec_experiments}).
Next, the connection to related work is discussed (Section~\ref{sec_related_work}).
Finally, we conclude and discuss possible future research directions (Section~\ref{sec_conclusion}).

% %%%%%%%%%%%%%%%%%%%%%%%%%%%%%%%%%%%%%%%%%%%%%%%%%% %
\section{Problem Statement and Notation}
\label{sec_prob_statement}
% %%%%%%%%%%%%%%%%%%%%%%%%%%%%%%%%%%%%%%%%%%%%%%%%%% %
Optimizing policies for \acp{MDP} with unknown dynamics is generally a hard problem (Section~\ref{sec_MDP}).
Specifically, this problem is hard due to the simulation optimization bias 
(Section~\ref{sec_SOB}), which is related to the optimality gap (Section~\ref{sec_OG}).
We derive an upper bound on the optimality gap, show its monotonic decrease with increasing number of samples from the random variable. Moreover, we clarify the relationship between the simulation optimization bias and the optimality gap (Section~\ref{sec_connection_OG_SOB}).
In what follows, we build upon the results of \cite{Hobbs_Hepenstal_1989,Mak_etal_1999}.

% -------------------------------------------------- %
\subsection{\acl{MDP}}
\label{sec_MDP}
% -------------------------------------------------- %
Consider a time-discrete dynamical system
\begin{equation}
\begin{gathered}
	\label{eq_sys_dyn}
	\fs_{t+1} \sim \transprob[\domparams]\left( \given{\fs_{t+1}}{\fs_t, \fa_t, \domparams} \right),\quad
	\fs_0 \sim \initstatedistr[,\domparams]( \given{\fs_0}{\domparams}), \\
	\fa_t \sim \pol{\given{\fa_t}{\fs_t; \polparams}},\quad
	\domparams \sim \domparamdistr{\domparams; \domdistrparams},
\end{gathered}
\end{equation}
with the continuous state ${\fs_t \in \stateset[\domparams] \subseteq \RR^{\dimstate}}$, and continuous action ${\fa_t \in \actionset[\domparams] \subseteq \RR^{\dimact}}$ \mbox{at time step $t$}. The environment, also called domain, is instantiated through its parameters ${\domparams \in \RR^{\dimdomparam}}$ (\eg, masses, friction coefficients, or time delays), which are assumed to be random variables distributed according to the probability distribution ${\domparamdistrsym \colon \RR^{\dimdomparam} \to \RR^{+}}$ parametrized by $\domdistrparams$.
These parameters determine the transition probability density function ${\transprob[\domparams] \colon \stateset[\domparams] \times \actionset[\domparams] \times \stateset[\domparams] \to \RR^{+}}$ that describes the system's stochastic dynamics.
The initial state $\fs_0$ is drawn from the start state distribution ${\initstatedistr[,\domparams] \colon \stateset[\domparams] \to \RR^{+}}$.
Together with the reward function ${\rewsym \colon \stateset[\domparams] \times \actionset[\domparams] \to \RR}$, and the temporal discount factor ${\gamma \in [0,1]}$, the system forms a \ac{MDP} described by the tuple ${\mdp[\domparams] = \left\langle \stateset[\domparams], \actionset[\domparams], \transprob[\domparams], \initstatedistr[,\domparams], r, \gamma \right\rangle}$.

% Expected discounted return
The goal of a \acf{RL} agent is to maximize the expected (discounted) return, a numeric scoring function which measures the policy's performance. The expected discounted return of a stochastic domain-independent policy $\pol{\given{\fa_t}{\fs_t; \polparams}}$, characterized by its parameters $\polparams \in \polparamspace \subseteq \RR^{\dimpolparam}$, is defined as
\begin{equation}
	\label{eq_edr}
	\edr{\polparams,\domparams,\fs_0} = \Esub{\traj}{\sum_{t=0}^{T-1} \gamma^t r(\fs_t, \fa_t) \Bigg| \polparams,\domparams, \fs_0}.
\end{equation}
While learning from trial and error, the agent adapts its policy parameters.
The resulting state-action-reward tuples are collected in trajectories, \aka rollouts, $\traj = \{\fs_t,\fa_t,r_t\}_{t=0}^{T-1}$, with ${r_t = r(\fs_t, \fa_t)}$.
To keep the notation concise, we omit the dependency on the initial state $\fs_0$.

% -------------------------------------------------- %
\subsection{\acf{SOB}}
\label{sec_SOB}
% -------------------------------------------------- %
\begin{figure}[t]
	\centering
	\def\svgwidth{0.85\columnwidth}
	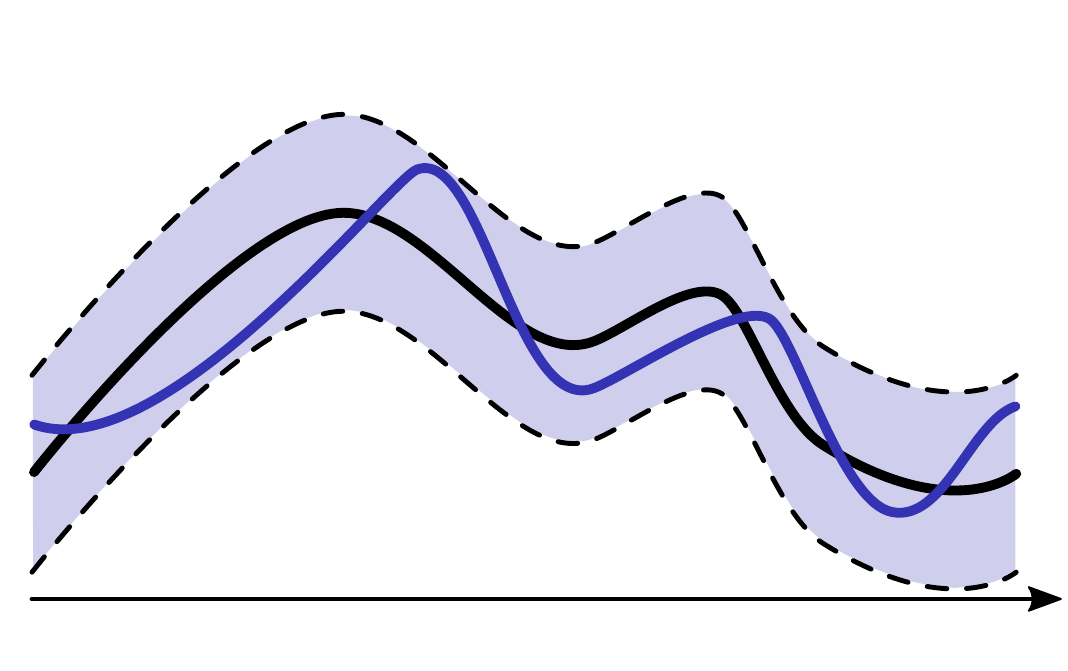
	\caption{%
		\acf{SOB} between the true optimum $\polparamopt$ and the sample-based optimum $\polparamopt[n]$. The shaded region visualizes the standard deviation around $\edr{\polparam}$, and $\estedr[n]{\polparam}$ is determined by a particular set of $n$ sampled domain parameters.
	}%
	\label{fig_SOB}
\end{figure}
\begin{table*}[t]
	\centering
	\caption{Definition and of the expectation of the expected (discounted) return, the \acf{SOB}, the \acf{OG}, and its estimation. All approximations are based on $n$ domains.}
	\renewcommand{\arraystretch}{2}
\rowcolors{2}{gray!25}{white}
\begin{tabular}[5pt]{lll}
 	\rowcolor{gray!50}
 	\textbf{Name} & \textbf{Definition} & \textbf{Property}\\
 	estimated expectation of the expected return & $\esteedr[n]{\polparams} = \frac{1}{n} \sum_{i=1}^{n} \edr{\polparams,\domparams_i}$ & $\Esub{\domparams}{\esteedr[n]{\polparams}} = \eedr{\polparams}$ \\
 	simulation optimization bias & $\sob{\estedr[n]{\polparamsopt[n]}} = \EsubBig{\domparams}{\max_{\hat{\polparams}\in\Theta} \estedrsym_n(\hat{\polparams})} - \max_{\polparams\in\Theta} \EsubBig{\domparams}{\edr{\polparams,\domparams}}$ & $\sob{\estedr[n]{\polparamsopt[n]}} \ge 0$ \\
 	optimality gap at solution $\solcandi$ & $\optgap{\solcandi} = \max_{\polparams\in\Theta} \Esub{\domparams}{\edr{\polparams,\domparams}} - \Esub{\domparams}{\edr{\solcandi,\domparams}}$ & $\optgap{\solcandi} \ge 0$ \\
 	estimated optimality gap at solution $\solcandi$ & $\estoptgap[n]{\solcandi} = \max_{\polparams\in\Theta} \esteedr[n]{\polparams} - \esteedr[n]{\solcandi}$ & $\estoptgap[n]{\solcandi} \ge \optgap{\solcandi}$ \\
\end{tabular}
	\label{tab_summary_notation}
\end{table*}

Augmenting the standard \ac{RL} setting with the concept of domain randomization, \ie maximizing the expectation of the expected return over all (feasible) realizations of the source domain, leads to the score
\begin{equation}
	\label{eq_eedr}
	\eedr{\polparams} = \Esub{\domparams}{\edr{\polparams,\domparams}}
\end{equation}
that quantifies how well the policy is expected to perform over an infinite set of variations of the nominal domain $\mdp[\domparamsnom]$.
When training exclusively in simulation, the true physics model is unknown and the true $\eedr{\polparams,\domparams}$ is thus inaccessible. Instead, we maximize the estimated expected return using a randomized physics simulator. Thereby, we update the policy parameters $\polparams$ with a policy optimization algorithm based on samples.
% Exploit of the imperfecions in simulation
The inevitable imperfections of physics simulations will automatically be exploited by any optimization method to achieve a `virtual' improvement, \ie, an increase of $\eedr{\polparams}$, in simulation.
% Stochasic Programs
To formulate this undesirable behavior, we frame the standard \ac{RL} problem as a \acf{SP}
\begin{equation}
	\label{eq_SP}
	\eedr{\polparamsopt} = \max_{\polparams\in\polparamspace} \Esub{\domparams}{\edr{\polparams,\domparams}} = \max_{\polparams\in\polparamspace} \eedr{\polparams},
\end{equation}
with the optimal solution $\polparamsopt = \argmax_{\polparams\in\polparamspace} \eedr{\polparams}$.

\noindent The \ac{SP} above can be approximated using $n$ domains
\begin{equation}
	\label{eq_SPn}
	\esteedr[n]{\polparamsopt[n]} = \max_{\polparams\in\polparamspace} \esteedr[n]{\polparams} = \max_{\polparams\in\polparamspace} \frac{1}{n} \sum_{i=1}^{n} \edr{\polparams,\domparams_i},
\end{equation}
where the expectation is replaced by the Monte-Carlo estimator over the samples $\domparams_1, \dots, \domparams_n$, and ${\polparamsopt[n] = \argmax_{\polparams\in\polparamspace} \esteedr[n]{\polparams}}$ is the solution to the approximated \ac{SP}.
Note that the expectations in (\ref{eq_SP}, \ref{eq_SPn}) both jointly depend on $\domparams$ and $\fs_0$, \ie both random varaibles are integrated out, but the dependency on $\fs_0$ is omitted as stated before.

% Simualtion optimization bias
Sample-based optimization is guaranteed to be optimistically biased if there are errors in the domain parameter estimate, even if these errors are unbiased~\cite{Hobbs_Hepenstal_1989}. Since the proposed method randomizes the domain parameters $\domparams$, this assumption is guaranteed to hold.
Using Jensen's inequality, we can show that the \acf{SOB}
\begin{equation}
\label{eq_SOB_def}
\sob{\estedr[n]{\polparamsopt[n]}} = \underbrace{\EsubBig{\domparams}{\max_{\hat{\polparams}\in\polparamspace} \estedrsym_n(\hat{\polparams})}}_{\text{sample optimum}} - \underbrace{\max_{\polparams\in\polparamspace} \EsubBig{\domparams}{\edr{\polparams,\domparams}}}_{\text{true optimum}} \ge 0.%,
\end{equation}
is always positive, \ie the policy's performance in the target domain is systematically overestimated.
%Regarding the \ac{SP} \eqref{eq_SP} and its approximation, we define the \acf{SOB} to be
%\begin{equation}
%	\label{eq_SOB_def}
%	\sob{\estedr[n]{\polparamsopt[n]}} = \underbrace{\EsubBig{\domparams}{\max_{\hat{\polparams}\in\polparamspace} \estedrsym_n(\hat{\polparams})}}_{\text{sample optimum}} - \underbrace{\max_{\polparams\in\polparamspace} \EsubBig{\domparams}{\edr{\polparams,\domparams}}}_{\text{true optimum}} \ge 0.%,
%\end{equation}
%The property ${\sobNormal{\estedr[n]{\polparamsopt[n]}} \ge 0}$ follows from the Jensen's inequality.
A visualization of the \ac{SOB} is depicted in Figure~\ref{fig_SOB}.

% -------------------------------------------------- %
\subsection{\acf{OG}}
\label{sec_OG}
% -------------------------------------------------- %
Intuitively, we want to minimize the \ac{SOB} in order to achieve the highest transferability of the policy.
Since computing the \ac{SOB} \eqref{eq_SOB_def} is intractable, the approach presented in this paper is to approximate the \acf{OG}, which relates to the \ac{SOB} as explained in the Section~\ref{sec_connection_OG_SOB}.

% Optimality gap definition
The \ac{OG} at the solution candidate $\solcandi$ is defined as
\begin{equation}
	\label{eq_optgap_def}
	\optgap{\solcandi} = \eedr{\polparamsopt} - \eedr{\solcandi} \ge 0,
\end{equation}
where ${\eedr{\polparamsopt} = \max_{\polparams\in\polparamspace} \Esub{\domparams}{\edr{\polparams,\domparams}}}$ is the \ac{SP}'s optimal objective function value and ${\eedr{\solcandi} = \Esub{\domparams}{\edr{\solcandi,\domparams}}}$ is the \ac{SP}'s objective function evaluated at the candidate solution \cite{Mak_etal_1999}. 
Thus, $\optgap{\solcandi}$ expresses the difference in performance between the optimal policy and the candidate solution at hand.
%When replacing the arbitrary $\solcandi$ with $\polparamsopt[n]$, \eqref{eq_optgap_def} reveals that $\polparamsopt[n]$ is a biased estimator of $\polparamsopt$.
Unfortunately, computing the expectation over infinitely many domains in \eqref{eq_optgap_def} is intractable. However, we can estimate $\optgap{\solcandi}$ from samples.

% -------------------------------------------------- %
\subsubsection{Estimation of the \acl{OG}}
\label{sec_OG_estimation}
% -------------------------------------------------- %
% Estimation of the optimality gap
For an unbiased estimator $\esteedr[n]{\polparams}$, \eg a sample average with \iid samples, he have
\begin{equation}
	\label{eq_Jn_unbiased_estitmator}
	 \Esubbig{\domparams}{\esteedr[n]{\polparams}} = \Esub{\domparams}{\edr{\polparams,\domparams}} = \eedr{\polparams}.
\end{equation}
Inserting \eqref{eq_Jn_unbiased_estitmator} into the first term of \eqref{eq_optgap_def} yields
\begin{align}
	\optgap{\solcandi} &= \max_{\polparams\in\polparamspace} \Esub{\domparams}{\esteedr[n]{\polparams}} - \Esub{\domparams}{\edr{\solcandi,\domparams}}\\
	&\le \Esub{\domparams}{\max_{\polparams\in\polparamspace} \esteedr[n]{\polparams}} - \Esub{\domparams}{\edr{\solcandi,\domparams}} \label{eq_optgap_final}
\end{align}
as an upper bound on the \ac{OG}. To compute this upper bound, we use the law of large numbers for the first term and replace the second expectation in \eqref{eq_optgap_final} with the sample average
\begin{equation}
	\optgap{\solcandi} \le \max_{\polparams\in\polparamspace} \esteedr[n]{\polparams} - \esteedr[n]{\solcandi} =  \estoptgap[n]{\solcandi}, \label{eq_optgap_approx}
\end{equation}
% Consistency of the result
where $ \estoptgap[n]{\solcandi} \ge 0$ holds.\footnote{This result is consistent with Theorem~1 and Equation~(9) in \cite{Mak_etal_1999} as well as the \enquote{type A error} in \cite{Hobbs_Hepenstal_1989}.}
Averaging over a finite set of domains allows for the utilization of an estimated upper bound of the \ac{OG} as the convergence criterion for the policy search meta-algorithm introduced in Section \ref{sec_SPOTA}.

% -------------------------------------------------- %
\subsubsection{Decrease of the Estimated \acl{OG}}
\label{sec_OG_decrease}
% -------------------------------------------------- %
The \ac{OG} decreases in expectation with increasing sample size of the domain parameters $\domparams$. 
The expectation over $\domparams$ of the minuend in \eqref{eq_optgap_approx} estimated from $n+1$ \iid samples is % written as
\begin{align}
	\Esub{\domparams}{\esteedr[n+1]{\polparamsopt[n+1]}}
	&= \Esub{\domparams}{\max_{\polparams\in\polparamspace} \frac{1}{n+1} \sum\limits_{i=1}^{n+1} \edr{\polparams,\domparams_i}}\\
	&= \Esub{\domparams}{\max_{\polparams\in\polparamspace} \frac{1}{n+1} \sum\limits_{i=1}^{n+1} \frac{1}{n} \sum\limits_{j=1,j\neq i}^{n+1} \edr{\polparams,\domparams_j}}\\
	&\le \Esub{\domparams}{\frac{1}{n+1} \sum\limits_{i=1}^{n+1} \max_{\polparams\in\polparamspace} \frac{1}{n} \sum\limits_{j=1,j\neq i}^{n+1} \edr{\polparams,\domparams_j}}\\
	%&= \frac{1}{n+1} \sum\limits_{i=1}^{n+1} \Esub{\domparams}{\max_{\polparams\in\polparamspace} \esteedr[n]{\polparams}}
	&= \Esub{\domparams}{\esteedr[n]{\polparamsopt[n]}}. \label{eq_optgap_approx_nplus1}
\end{align}
Taking the expectation of the \ac{OG} estimated from $n+1$ samples $\estoptgap[n+1]{\solcandi}$ and then plugging in the upper bound from \eqref{eq_optgap_approx_nplus1}, we obtain the upper bound
%%\begin{equation}
%%	\Esub{\domparams}{\estoptgap[n+1]{\solcandi}} \le
%%	\mathbb{E}_{\domparams} \bigg[ \underbrace{\max_{\polparams\in\polparamspace} \esteedr[n]{\polparams} - \Esub{\domparams}{\edr{\solcandi,\domparams}}}_{\estoptgap[n]{\solcandi}} \bigg]
%%	%\Esub{\domparams}{\underbrace{\max_{\polparams\in\polparamspace} \esteedr[n]{\polparams} - \Esub{\domparams}{\edr{\solcandi,\domparams}}}_{\estoptgap[n]{\solcandi}}},
%%\end{equation}
\begin{align}
	\Esub{\domparams}{\estoptgap[n+1]{\solcandi}} &= \Esub{\domparams}{\max_{\polparams\in\polparamspace} \esteedr[n+1]{\polparams} - \esteedr[n]{\solcandi}}\\
	&\le \Esub{\domparams}{\max_{\polparams\in\polparamspace} \esteedr[n]{\polparams} - \esteedr[n]{\solcandi}} = \Esub{\domparams}{\estoptgap[n]{\solcandi}},
\end{align}
which shows that the estimator of the \ac{OG} in expectation monotonically decreases with increasing sample size.\footnote{This result is consistent with Theorem~2 in~\cite{Mak_etal_1999}.}

% -------------------------------------------------- %
\subsection{Connection Between the \ac{SOB} and the \ac{OG}}
\label{sec_connection_OG_SOB}
% -------------------------------------------------- %
The \ac{SOB} can be expressed as the expectation of the difference between the approximated \ac{OG} and the true \ac{OG}. 
Starting from the formulation of the approximated \ac{OG} in \eqref{eq_optgap_approx}, we can take the expectation over the domains on both sides of the inequality and rearrange to
\begin{equation}
	\Esub{\domparams}{\estoptgap[n]{\solcandi}} - \optgap{\solcandi} \ge 0.
\end{equation}
Using the definitions of $\estoptgap[n]{\solcandi}$ and $\optgap{\solcandi}$ from Table~\ref{tab_summary_notation}, the equation above can be rewritten as
%\begin{align}
%	&\Esub{\domparams}{\max_{\polparams\in\polparamspace} \esteedr[n]{\polparams} - \esteedr[n]{\solcandi}} - \\
%	&\indent\max_{\polparams\in\polparamspace} \Esub{\domparams}{\edr{\polparams,\domparams}} - \Esub{\domparams}{\edr{\solcandi,\domparams}} \ge 0,
%\end{align}
%which simplifies to
\begin{align}
	&\Esub{\domparams}{\max_{\polparams\in\polparamspace} \esteedr[n]{\polparams}} - \Esub{\domparams}{\esteedr[n]{\solcandi}} - \\
	&\indent\max_{\polparams\in\polparamspace} \Esub{\domparams}{\edr{\polparams,\domparams}} + \EsubBig{\domparams}{\edr{\solcandi,\domparams}} \ge 0. \label{eq_ogap_sob_interm}
\end{align}
Since $\esteedr[n]{\polparams}$ is an unbiased estimator of $\edr{\polparams}$, we have %it is
%\begin{align}
%\hspace{-2em}
%\Esub{\domparams}{\esteedr[n]{\solcandi}} &= \Esub{\domparams}{\frac{1}{n} \sum_{i=1}^{n} \edr{\solcandi, \domparams_i}}\\
%&= \frac{1}{n} \sum_{i=1}^{n} \Esub{\domparams}{\edr{\solcandi, \domparams}} = \Esub{\domparams}{\edr{\solcandi, \domparams}}. \label{eq_Eestret_equal_Eret}
%\end{align}
\begin{equation}
	\label{eq_Eestret_equal_Eret}
	\Esub{\domparams}{\esteedr[n]{\solcandi}} = \Esub{\domparams}{\edr{\solcandi, \domparams}} = \eedr{\solcandi}.
\end{equation}
Hence, the left hand side of \eqref{eq_ogap_sob_interm} becomes
\begin{equation}
	\Esub{\domparams}{\max_{\polparams\in\polparamspace} \esteedr[n]{\polparams}} -
	\max_{\polparams\in\polparamspace} \EsubBig{\domparams}{\edr{\polparams,\domparams}} = \sob{\estedr[n]{\polparamsopt[n]}},
\end{equation}
which is equal to the \ac{SOB} defined in \eqref{eq_SOB_def}. Thus, the \ac{SOB} is the difference between the expectation over all domains of the estimated \ac{OG} $\estoptgap[n]{\solcandi}$ and the true \ac{OG} $\optgap{\solcandi}$ at the solution candidate.
Therefore, reducing the estimated \ac{OG} leads to reducing the \ac{SOB}.

% -------------------------------------------------- %
\subsection{An Illustrative Example}
% -------------------------------------------------- %
Imagine we were placed randomly in an environment either on Mars $\domparams_M$ or on Venus $\domparams_V$, governed by the distribution $\domparams \sim \domparamdistr{\domparams; \domdistrparam}$. On both planets we are in a catapult about to be shot into the sky exactly vertical. The only thing we can do is to manipulate the catapult, modeled as a linear spring, according to the policy $\pol{\polparam}$, \ie changing the springs extension.
Our goal is to minimize the maximum height of the expected flight trajectory $\Esub{\domparams}{h(\polparam, \domparams)}$ derived from the conservation of energy
\begin{equation}
	h(\polparam, \domparams_i) = \frac{k_i (\polparam - x_i)^2}{2 m g_i},
\end{equation}
with mass $m$, and domain parameters ${\domparams_i = \{g_i, k_i, x_i\}}$ consisting of the gravity acceleration constant, the catapult's spring stiffness, and the catapult's spring pre-extension. The domain parameters are the only quantities specific to Mars and Venus.
In this simplified example, we assume that the domain parameters are not drawn from individual distributions, but that there are two sets of domain parameters $\domparams_M$ and $\domparams_V$ which are drawn from a Bernoulli distribution $\distrbern{\domparams}{\domdistrparam}$ where $\domdistrparam$ is the probability of drawing $\domparams_V$.
%\\
Since minimizing $\Esub{\domparams}{h(\polparam, \domparams)}$ is identical to maximizing its negative value $\edr{\polparam, \domparams} \coloneqq -\Esub{\domparams}{h(\polparam, \domparams)}$, we rewrite the problem as
\begin{equation}
	\eedr{\polparamopt} = \max_{\polparam\in\polparamspace} \Esub{\domparams}{\edr{\polparam,\domparams}}.
\end{equation}
Assume we experienced this situation $n$ times and want to find the policy parameter maximizing the objective above without knowing on which planet we are (\i.e., independent of $\domparams$). Thus, we approximate $\eedr{\polparamopt}$ by
\begin{equation}
	\esteedr[n]{\polparamopt[n]} = \max_{\polparam\in\polparamspace} \frac{1}{n} \sum_{i=1}^{n} \edr{\polparam, \domparams_i}.
\end{equation}
In this Bernoulli experiment, the return of a policy $\pol{\polparam}$ fully determined by $\polparam$ estimates to
%\begin{align}
%\esteedr[n]{\polparam} &= \underbrace{\frac{n_M}{n} \edr{\polparam, \domparams_M}}_{\text{proportion Mars}} + \underbrace{\frac{n_V}{n} \edr{\polparam, \domparams_V}}_{\text{proportion Venus}}\\
%	&= -\frac{n_M}{n} \frac{k_M (\polparam - x_M)^2}{2 m g_M} -\frac{n_V}{n} \frac{k_V (\polparam - x_V)^2}{2 m g_V} \label{eq_ill_exmpl_est_eedr}.
%\end{align}
\begin{equation}
	\label{eq_ill_exmpl_est_eedr}
	\esteedr[n]{\polparam} = \underbrace{\frac{n_M}{n} \edr{\polparam, \domparams_M}}_{\text{proportion Mars}} + \underbrace{\frac{n_V}{n} \edr{\polparam, \domparams_V}}_{\text{proportion Venus}}.
\end{equation}
The optimal policy given the $n$ domains fulfills the necessary condition
\begin{equation}
	0 = \nabla_\polparam \esteedr[n]{\polparamopt[n]} =
	-\frac{n_M}{n} \frac{k_M (\polparamopt[n] - x_M)}{m g_M} -\frac{n_V}{n} \frac{k_V (\polparamopt[n] - x_V)}{m g_V}.
%	\\0 &= n_M k_M g_V (\polparamopt[n] - x_M) + n_V k_V g_M (\polparamopt[n] - x_V)
\end{equation}
Solving for the optimal policy parameter yields
\begin{equation}
	\hspace{-0.19em}\polparamopt[n]
	= \frac{ x_M n_M k_M g_V + x_V n_V k_V g_M }{ n_M k_M g_V + n_V k_V g_M }
	= \frac{x_M c_M + x_V c_V}{c_M + c_V},
	\label{eq_ill_exmpl_polparamopt}
\end{equation}
with the (mixed-domain) constants ${c_M =  n_M k_M g_V}$ and ${c_V = n_V k_V g_M}$.
Inserting \eqref{eq_ill_exmpl_polparamopt} into \eqref{eq_ill_exmpl_est_eedr} gives the optimal return value for $n$ samples
\begin{align}
	\esteedr[n]{\polparamopt}
%	&= -\frac{\frac{n_M}{n} k_M}{2 m g_M} \left(\frac{x_M c_M + x_V c_V}{c_M + c_V} - x_M \right)^2\\
%	&\qquad -\frac{\frac{n_V}{n} k_V}{2 m g_V} \left(\frac{x_M c_M + x_V c_V}{c_M + c_V} - x_V \right)^2\\
	&= -\frac{n_M k_M}{2 n m g_M} \left(\frac{x_V c_V - x_M c_V}{c_M + c_V} \right)^2\\
	&\qquad -\frac{n_V k_V}{2 n m g_V} \left(\frac{x_M c_M - x_V c_M}{c_M + c_V} \right)^2.
\end{align}
% Solving the example for given values
Given the domain parameters in Table~\ref{tab_params_ill_exmpl} of Appendix~\ref{secapdx_hparams}, we optimize our catapult manipulation policy. This is done in simulation, since real-world trials (being shot with a catapult) would be very costly.
Finding the optimal policy in simulation means solving the approximated \ac{SP} \eqref{eq_SPn}, whose optimal solution is denoted by $\polparamopt[n]$.
We assume that the (stochastic) optimization algorithm outputs a suboptimal solution $\polparam^c$. In order to model this property, a policy parameter is sampled in the vicinity of the optimum ${\polparam^c \sim \distrnormal{\polparam}{\polparamopt[n]; \sigma^2_\polparam}}$ with ${\sigma_\polparam = 0.15}$.
During the entire process, the true optimal policy parameter $\polparamopt$ will remain unknown. However, since this simplified example models the domain parameters to be one of two fixed sets ($\domparams_M$ or $\domparams_V$), $\polparamopt$ can be computed analogously to \eqref{eq_ill_exmpl_polparamopt}.

% Key information from figures
	\begin{figure}[t]
		\centering
		\includegraphics[height=5.0cm,keepaspectratio]{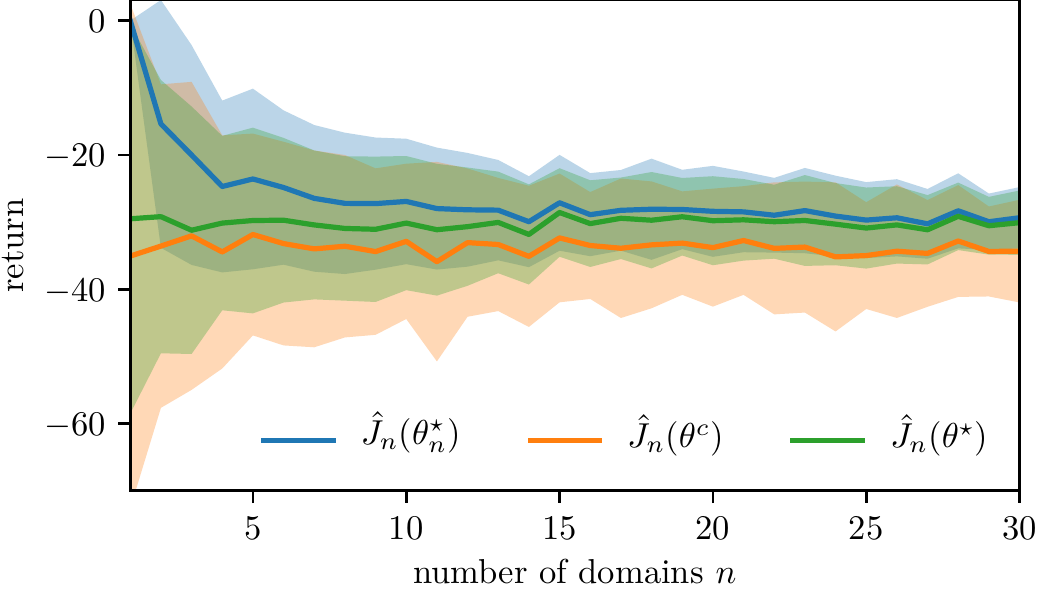}%[width=\linewidth,keepaspectratio] or [width=\columnwidth]
		\caption{%
			The estimated expected return evaluated using the optimal solution for a set of $n$ domains $\estedr[n]{\polparamopt[n]}$, the candidate solution $\estedr[n]{\polparam^c}$, as well as the true optimal solution $\estedr[n]{\polparamopt}$. Note that $\esteedr[n]{\polparamopt[n]} > \esteedr[n]{\polparam^c}$ holds for every instance of the 100 random seeds, even if the standard deviation areas overlap.
			The shaded areas show $\pm 1$ standard deviation.
		}
		\label{fig_ill_exmpl_Jn}
	\end{figure}
\begin{figure}[t]
	\centering
	\includegraphics[height=5.0cm,keepaspectratio]{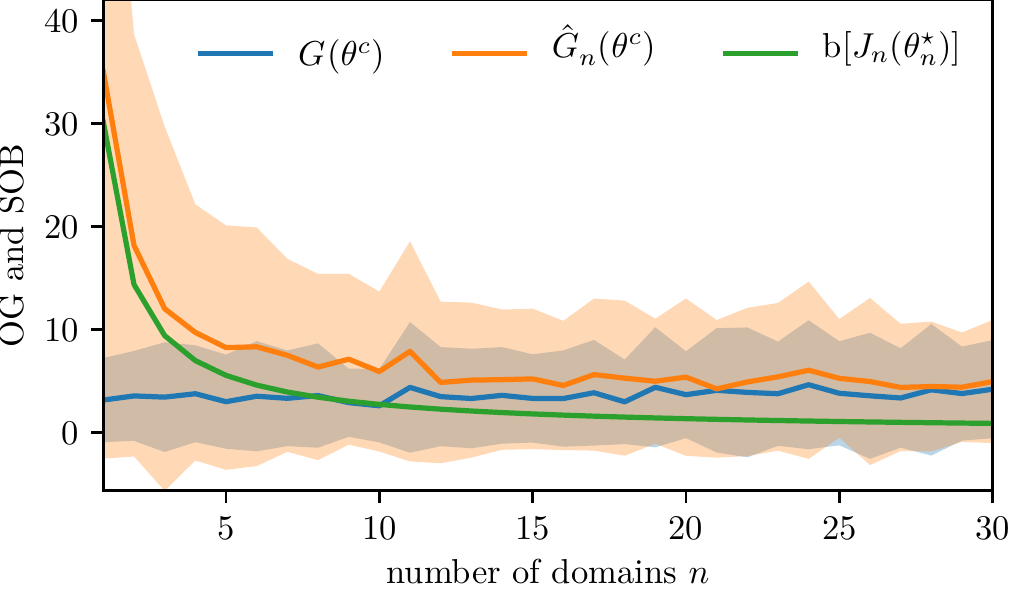}%[width=\linewidth,keepaspectratio] or [width=\columnwidth]
	\caption{%
		True optimality gap $\optgap{\polparam^c}$, the approximation from $n$ domains $\estoptgap{\polparam^c}$, and the simulation optimization bias $\sobNormal{\estedr[n]{\polparamopt[n]}}$. 
		Note that $\estoptgap[n]{\polparam^c} \ge \optgap{\polparam^c}$ does not hold for every instance of the 100 random seeds, but is true in expectation.
		The variance in $\optgap{\polparam^c}$ is only caused by the variance in $\polparam^c$.
		The shaded areas show $\pm 1$ standard deviation.
	}
	\label{fig_ill_exmpl_OG_SOB}
\end{figure}

The Figures~\ref{fig_ill_exmpl_Jn}~and~\ref{fig_ill_exmpl_OG_SOB} visualize the evolution of the approximated \ac{SP} with increasing number of domains $n$. Key observations are that %$\polparamopt[n]$ converges to $\polparamopt$ (Figure~\ref{fig_ill_exmpl_theta})
the objective function value at the candidate solution $\esteedr[n]{\polparam^c}$ is less than at the sample-based optimum $\esteedr[n]{\polparamopt[n]}$ (Figure~\ref{fig_ill_exmpl_Jn}), and that with increasing number of domains the \ac{SOB} $\bias{\estedr{\polparamopt[n]}}$ decreases monotonically while the estimated \ac{OG} $\estoptgap[n]{\polparam^c}$ only decreases in expectation (Figure~\ref{fig_ill_exmpl_OG_SOB}).
% Intering final values
%In terms of the catapult example, the interpretation of the final values in Figure~\ref{fig_ill_exmpl_OG_SOB} is as follows.
When optimizing over $n = 30$ random domains in simulation, we yield a policy which leads to
%\begin{itemize}
%	\item
a $\optgap[n]{\polparam^c} \approx \SI{4.23}{\meter}$ higher (worse) shot compared to the best policy computed from an infinite set of domains and evaluated on this infinite set of domains,
and a $\estoptgap[n]{\polparam^c} \approx \SI{4.97}{\meter}$ higher (worse) shot compared to the best policy computed from a set of $n = 30$ domains and evaluated on the same finite set of domains.
%\end{itemize}
Furthermore, we can say that executing the best policy computed from a set of $n = 30$ domains will in reality result in a ${\bias{\estedr{\polparamopt[n]}} \approx \SI{0.911}{\meter}}$ higher (worse) shot.

% %%%%%%%%%%%%%%%%%%%%%%%%%%%%%%%%%%%%%%%%%%%%%%%%%% %
\section{Simulation-Based Policy Optimization with Transferability Assessment}
\label{sec_SPOTA}
% %%%%%%%%%%%%%%%%%%%%%%%%%%%%%%%%%%%%%%%%%%%%%%%%%% %
% Abstract of SPOTA
We introduce \acf{SPOTA}~\cite{Muratore_etal_2018}, a policy search meta-algorithm which yields a policy that is able to directly transfer from a set of source domains to an unseen target domain.
% Goal and key novelty of SPOTA
The goal of \ac{SPOTA} is not only to maximize the expected discounted return under the influence of randomized physics simulations $\eedr{\polparams}$, but also to provide an approximate probabilistic guarantee on the suboptimality in terms of expected discounted return when applying the obtained policy to a different domain.
The key novelty in \ac{SPOTA} is the utilization of an \acf{UCBOG} as a stopping criterion for the training procedure of the \ac{RL} agent.

% Domain randomization as uncertainty representation
One interpretation of (source) domain randomization is to see it as a form of uncertainty representation. If a control policy is trained successfully on multiple variations of the scenario, \ie , a set of models, it is legitimate to assume that this policy will be able to handle modeling errors better than policies that have only been trained on the nominal model $\domparamsnom = \Esub{\domparamdistrsym}{\domparams}$.
With this rationale in mind, we propose the \acs{SPOTA} procedure, summarized in Algorithm~\ref{algo_SPOTA}.

%\begin{tikzpicture}
%\begin{axis}[
%scale=0.5
%]
%\input{plots/illustrative_example/OG_SOB.pgf}
%\end{axis}
%\end{tikzpicture}

% -------------------------------------------------- %
\begin{algorithm*}[t] % [H] no floating behavior in IEEE template
	\caption{\acf{SPOTA}} % caption requires algorithm environment
	\label{algo_SPOTA}
	\DontPrintSemicolon
\Input{probability distribution $\domparamdistr{\domparams;\domdistrparams}$, algorithm \texttt{PolOpt}, sequence \texttt{NonDecrSeq}, \mbox{hyper-parameters $n_c$, $\refsym$, $n_G$, $n_J$, $\numbssamples$, $\alpha$, $\beta$}}
\Output{policy $\pol{\polparamsoptcand}$ with a $(1-\alpha)$-level confidence on $\optgapmean[\refsym]{\polparamsoptcand}$ which is upper bounded by $\beta$}

Initialize $\pol{\polparams[n_c]}$ randomly \label{SPOTA_init}\;
% Start do-while loop
\Do{$\optgapmean[\refsym][U]{\polparamsoptcand} > \beta$}
{
	% candidate solution
	Sample $\candsym$ \iid physics simulators described by $\domparams_1, \dots, \domparams_{\candsym}$ from $\domparamdistr{\domparams;\domdistrparams}$ \label{SPOTA_cand_DR}\; % realizations ... \Comment*[r]{$\candsym$ different phys. simulators}
	Solve the approx. \ac{SP} using $\domparams_1, \dots, \domparams_{\candsym}$ and \texttt{PolOpt} to obtain $\polparamsoptcand$ \label{SPOTA_optpol_cand} \Comment*[r]{candidate solution}
	\For{$k = 1,\dots,n_G$}
	{
		% reference solution(s)
		Sample $\refsym$ \iid physics simulators described by $\domparams_1^k,\dots,\domparams_{\refsym}^k$ from $\domparamdistr{\domparams;\domdistrparams}$ \label{SPOTA_ref_DR}\; % realizations ... \Comment*[r]{$\refsym$ different phys. simulators}
		Initialize $\polparams[\refsym][k]$ with $\polparamsoptcand$ and reset the exploration strategy\label{SPOTA_ref_init}\;
		Solve the approx. \ac{SP} using $\domparams_1^k,\dots,\domparams_{\refsym}^k$ and \texttt{PolOpt} to obtain $\polparamsopt[\refsym][k]$ \label{SPOTA_optpol_ref} \Comment*[r]{reference solution}
		\For{$i = 1,\dots,\refsym$}
		{
			\With(synchronized random seeds\label{SPOTA_seed_sync_start} \Comment*[f]{sync initial states and observation noise})
			{
				Estimate the candidate solution's return % for the $i$th realization
				$\estedr[n_J]{\polparamsoptcand,\domparams_i^k} \gets 1/n_J \sum_{j=1}^{n_J} \estedr{\polparamsoptcand,\domparams_i^k}$\;
				Estimate the $i$-th reference solution's return % for the $i$th realization
				$\estedr[n_J]{\polparamsopt[\refsym][k],\domparams_i^k} \gets 1/n_J \sum_{j=1}^{n_J} \estedr{\polparamsopt[\refsym][k],\domparams_i^k}$
			}\label{SPOTA_seed_sync_end}
			Compute the difference in return % for the $i$th realization
			 $\estoptgap[\refsym,i][k]{\polparamsoptcand} \gets \estedr[n_J]{\polparamsopt[\refsym][k],\domparams_i^k} - \estedr[n_J]{\polparamsoptcand,\domparams_i^k}$ \label{SPOTA_optgap_diff_sample}\;
		}
		\For(\Comment*[f]{outlier rejection}){$k = 1,\dots,n_G$ \kwAnd $i = 1,\dots,\refsym$}
		{\label{SPOTA_outlier_rejection_start}
			\If{$\estoptgap[\refsym,i][k]{\polparamsoptcand} < 0$}
			{
				\For(\Comment*[f]{loop over other reference solutions}){$k^\prime = 1,\dots,n_G$, $k^\prime \neq k$}
				{
					\llIf{$\estoptgap[\refsym,i][k^\prime]{\polparamsoptcand} > \estoptgap[\refsym,i][k]{\polparamsoptcand}$}
					{$\estoptgap[\refsym,i][k]{\polparamsoptcand} \gets \estoptgap[\refsym,i][k^\prime]{\polparamsoptcand}$; \kwBreak \Comment*[r]{replace solution}}
				}
%				\llIf{\kwNot replaced}{$\estoptgap[\refsym,i][k]{\polparamsoptcand} \gets 0$}
			}
		}\label{SPOTA_outlier_rejection_end}
	}
	Bootstrap $\numbssamples$ times from $\optgapsampset = \{\estoptgap[\refsym,1][1]{\polparamsoptcand}, \dots, \estoptgap[\refsym,\refsym][n_G]{\polparamsoptcand}\}$ to yield $\bootoptgapsampset[1], \dots, \bootoptgapsampset[\numbssamples]$ \Comment*[r]{bootstrapping}
	
	Compute the sample mean $\optgapmean[\refsym]{\polparamsoptcand}$ for the original set $\optgapsampset$ \label{SPOTA_mean_orig_samples}\;
	Compute the sample means $\bootmeanoptgap[\refsym,1]{\polparamsoptcand}, \dots , \bootmeanoptgap[\refsym,\numbssamples]{\polparamsoptcand}$ for the sets $\bootoptgapsampset[1],\dots,\bootoptgapsampset[\numbssamples]$ \label{SPOTA_mean_bs_samples}\;
	
	Select the $\alpha$-th quantile of the bootstrap samples' means and obtain the upper bound for the one-sided $(1-\alpha)$-level confidence interval $\optgapmean[\refsym][U]{\polparamsoptcand} \gets 2 \optgapmean[\refsym]{\polparamsoptcand} - \quantile{\alpha}{\bootmeanoptgap[\refsym]{\polparamsoptcand}}$ \Comment*[r]{\acs{UCBOG}} \label{SPOTA_UCBOG}
		
	Set the new sample sizes $n_c \gets \texttt{NonDecrSeq}(n_c)$ and $\refsym \gets \texttt{NonDecrSeq}(\refsym)$
}\label{SPOTA_stopping_criterion}

\end{algorithm*}
% -------------------------------------------------- %

% Super short abstract of SPOTA
%On the most abstract level, \ac{SPOTA} can be seen as a repetitive comparison of solution candidates in unknown domains against reference solutions trained exclusively in these domains.
\ac{SPOTA} performs a repetitive comparison of solution candidates against reference solutions in domains that are in the references' training set but unknown to the candidates.
% Algorithm description
% Init
As inputs, we assume a probability distribution over the domain parameters $\domparamdistr{\domparams;\domdistrparams}$, a policy optimization sub-routine \texttt{PolOpt}, the number of candidate and reference domains $\candsym$ and $\refsym$ in conjunction with a nondecreasing sequence \texttt{NonDecrSeq} (\eg, $n_{k+1} = 2 n_k$), the number of reference solutions $n_G$, the number of rollouts used for each \ac{OG} estimate $n_J$, the number of bootstrap samples $\numbssamples$, the confidence level (1$-\alpha$) used for bootstrapping, and the threshold of trust $\beta$ determining the stopping condition.
% Def of phases
\ac{SPOTA} consists of four blocks:
finding a candidate solution,
finding multiple reference solutions,
comparing the candidate against the reference solutions, and
assessing the candidate solution quality.

% -------------------------------------------------- %
\textbf{Candidate Solutions}
are randomly initialized and optimized based on a set of $\candsym$ source domains (Lines~\ref{SPOTA_cand_DR}--\ref{SPOTA_optpol_cand}).
Practically, the locally optimal policy parameters are optimized on the approximated \ac{SP} \eqref{eq_SPn}.

% -------------------------------------------------- %
\textbf{Reference Solutions}
are gathered $n_G$ times by solving the same approximated \ac{SP} as for the candidate but with different realizations of the random variable $\domparams$ (Lines~\ref{SPOTA_ref_DR}--\ref{SPOTA_optpol_ref}). These $n_G$ non-convex optimization processes all use the same candidate solution $\polparamsoptcand$ as initial guess. % (Line~\ref{SPOTA_ref_init}).

% -------------------------------------------------- %
\textbf{Solution Comparison}
is done by evaluating each reference solution $\polparamsopt[\refsym][k]$ with $k=1,\dots,n_G$ against the candidate solution $\polparamsoptcand$ for each realization of the random variable $\domparams_i^k$ with $i=1,\dots,\refsym$ on which the reference solution has been trained. 
%Each reference solution $\polparamsopt[\refsym][k]$ gets evaluated against the candidate solution $\polparamsoptcand$ for each realization of the random variable $\domparams[1][k],\dots,\domparams[\refsym][k]$ on which the reference solution has been trained. % (Line~\ref{SPOTA_optgap_diff_sample}).
In this step, the performances per domain
$\estedr[n_J]{\polparamsoptcand,\domparams_i^k}$ and $\estedr[n_J]{\polparamsopt[\refsym][k],\domparams_i^k}$
are estimated from $n_J$ Monte-Carlo simulations with synchronized random seeds (Lines~\ref{SPOTA_seed_sync_start}--\ref{SPOTA_seed_sync_end}).
%\begin{align}
%	\estedr[n_J]{\polparamsoptcand,\domparams_i^k} &= \sum_{j=1}^{n_J} \estedr[j]{\polparamsoptcand,\domparams_i^k}, \label{eq_est_cand_ret}\\
%	\estedr[n_J]{\polparamsopt[\refsym][k],\domparams_i^k} &= \sum_{j=1}^{n_J} \estedr[j]{\polparamsopt[\refsym][k],\domparams_i^k} \label{eq_est_ref_ret}.
%\end{align}
Thereby, both solutions are evaluated using the same random initial states and observation noise.
Due to the potential suboptimality of the reference solutions, the resulting difference in performance 
\begin{equation}
\label{eq_optgap_sample}
\estoptgap[\refsym,i][k]{\polparamsoptcand} = \estedr[n_J]{\polparamsopt[\refsym][k],\domparams_i^k} - \estedr[n_J]{\polparamsoptcand,\domparams_i^k} % or formulated in sums \sum_{j=1}^{n_J} \estedr[j]{\polparamsoptcand,\domparams_i^k} - \sum_{j=1}^{n_J} \estedr[j]{\polparamsoptcand,\domparams_i^k}
\end{equation}
may become negative (Line~\ref{SPOTA_optgap_diff_sample}).
% Negative optimality gap samples
This issue did not appear in previous work on assessing solution qualities of \acp{SP} \cite{Mak_etal_1999,Bayraksan_Morton_2006}, because they only covered convex problems, where all reference solutions are guaranteed to be global optima.
Utilizing the definition of the \ac{OG} in \eqref{eq_optgap_approx} for \ac{SPOTA} demands for globally optimal reference solutions. Due to the non-convexity of the introduced \ac{RL} problem the obtained solutions by the optimizer only are locally optimal.
In order to alleviate this dilemma, we perform an outlier rejection routine (Lines~\ref{SPOTA_outlier_rejection_start}--\ref{SPOTA_outlier_rejection_end}). As a first attempt, all other reference solutions are evaluated for the current domain $i$. If a solution with higher performance was found, it replaces the current reference solution $k$ for this domain. If all reference solutions are worse than the candidate, the value is clipped to the theoretical minimum (zero).

% -------------------------------------------------- %
\textbf{Solution Quality}
is assessed by constructing a (1$-\alpha$)-level confidence interval $\left[ 0, \optgapmean[\refsym][U]{\polparamsoptcand}\right]$ for the estimated \ac{OG} at $\polparamsoptcand$. While the lower bound is fixed to the theoretical minimum, the \acf{UCBOG} is computed using the statistical bootstrap method~\cite{Efron_1979}.
We denote bootstrapped quantities with the superscript $B$ instead of the common asterisk, to avoid a notation conflict with the optimal solutions.
There are multiple ways to yield a confidence interval by applying the bootstrap \cite{DiCiccio_Efron_1996}.
Here, the 'basic' nonparametric method was chosen, since the aforementioned potential clipping changes the distribution of the samples and hence a method relying on the estimation of population parameters such as the standard error is inappropriate.
% \\
The solution comparison yields a set of $n_G \refsym$ samples of the approximated \ac{OG} $\optgapsampset = \{\estoptgap[\refsym,1][1]{\polparamsoptcand}, \dots, \estoptgap[\refsym,\refsym][n_G]{\polparamsoptcand}\}$.
Through uniform random sampling with replacement from $\optgapsampset$, we generate $\numbssamples$ bootstrap samples $\bootoptgapsampset[1],\dots,\bootoptgapsampset[\numbssamples]$.
Thus, for our statistic of interest, the mean estimated \ac{OG} $\optgapmean[\refsym]{\polparamsoptcand}$, the \ac{UCBOG} becomes
\begin{equation}
	\label{eq_UCBOG}
	\optgapmean[\refsym][U]{\polparamsoptcand} = 2 \optgapmean[\refsym]{\polparamsoptcand} - \quantile{\alpha}{\bootmeanoptgap[\refsym]{\polparamsoptcand}},
\end{equation}
where $\optgapmean[\refsym]{\polparamsoptcand}$ is the mean over all (nonnegative) samples from the empirical distribution, and $\quantile{\alpha}{\bootmeanoptgap[\refsym]{\polparamsoptcand}}$ is the $\alpha$-th quantile of the means calculated for each of the $\numbssamples$ bootstrap samples (Lines~\ref{SPOTA_mean_orig_samples}--\ref{SPOTA_UCBOG}).
% Probability of being inside the CI
Consequently, the true \ac{OG} is lower than the obtained one-sided confidence interval with the approximate probability of (1$-\alpha$), \ie,
%${\Pr{\optgap{\polparamsoptcand} \le \optgapmean[\refsym][U]{\polparamsoptcand}} \approx 1-\alpha}$,
\begin{equation}
\Pr{\optgap{\polparamsoptcand} \le \optgapmean[\refsym][U]{\polparamsoptcand}} \approx 1-\alpha,
\end{equation}
which is analogous to (4) in \cite{Bayraksan_Morton_2006}.
% Increasing sample sizes
Finally, the sample sizes $\refsym$ and $\candsym$ of the next epoch are set according to the nondecreasing sequence.
% Stopping criterion
The procedure stops if the \ac{UCBOG} at $\polparamsoptcand$ is less than or equal to the specified threshold of trust $\beta$. Fulfilling this condition, the candidate solution at hand does not lose more than $\beta$ in terms of performance with approximate probability (1$-\alpha$), when it is applied to a different domain sampled from the same distribution.

% Difference of SPOTA to the related work
% Why not use all samples for training one policy?
Intuitively, the question arises why we do not use all samples for training a single policy and thus most likely yield a more robust result.
To answer this question we want to point out that the key difference of \ac{SPOTA} to the related methods is the assessment of the solution's transferability to different domains. While the approaches reviewed in Section~\ref{sec_related_work} train one policy until convergence (\eg, for a fixed number of steps)%or until the change of the expected return is below some threshold)
, \ac{SPOTA} repeats this process and suggests new policies as long as the \ac{UCBOG} is above a specified threshold.
Thereby, \ac{SPOTA} only uses $1/(1 + n_G n / n_c)$ of the total samples to learn the candidate solution, \ie, the policy which will be deployed. If we would use all samples for training, hence not learn any reference solutions, we would not be able to estimate the \ac{OG} and therefore lose the main feature of \ac{SPOTA}.

% %%%%%%%%%%%%%%%%%%%%%%%%%%%%%%%%%%%%%%%%%%%%%%%%%% %
\section{Experiments}
\label{sec_experiments}
% %%%%%%%%%%%%%%%%%%%%%%%%%%%%%%%%%%%%%%%%%%%%%%%%%% %
% Experiments overview
We evaluate \ac{SPOTA} on two \simtoreal tasks pictured in Figure~\ref{fig_Quanser_platforms}, the Ball-Balancer and the Cart-Pole.
The policies obtained by \ac{SPOTA} are compared against \ac{EPOpt}, and (plain) \ac{PPO} policies. %, as well as hand-tuned platform-specific controllers (\acs{PD}-controller for Ball-Balancer and energy-based controller for Cart-Pole).
% Goal of the experiments
The goal of the conducted experiments is twofold.
First, we want to investigate the applicability of the \ac{UCBOG} as a quantitative measure of a policy's transferability.
Second, we aim to show that domain randomization enables the \simtoreal transfer of control policies learned by \ac{RL} algorithms, while methods which only learn from the nominal domain fail to transfer.
%Note that for all real-world experiments we expect the classical controllers to outperform the \ac{RL}-based policies, because we tuned them by hand on the physical platforms. Thus, they serve as a reference.

% -------------------------------------------------- %
\subsection{Modeling and Setup Description}
\label{sec_modeling_setup_description}
% -------------------------------------------------- %
Both platforms can be classified as nonlinear under-actuated balancing problems with continuous state and action spaces.
% Ball-Balancer description
The Ball-Balancer's task is to stabilize the randomly initialized ball at the plate's center. Given measurements and their first derivatives (obtained by temporal filtering) of the motor shaft angles as well as the ball's position relative to the plate's center, the agent controls two servo motors via voltage commands.
The rotation of the motor shafts leads, through a kinematic chain, to a change in the plate angles. Finally, the plate's deflection gets the ball rolling. The Ball-Balancer has an 8D state and a 2D action space.
% Cart-Pole description
Similarly, the Cart-Pole's task is to stabilize a pole in the upright position by controlling the cart.
Based on measurements and their first derivatives (obtained by temporal filtering) of the pole's rotation angle as well as the cart position relative to the rail's center, the agent controls the servo motor driving the cart by sending voltage commands. Accelerating the cart makes the pole rotate around an axis perpendicular to the cart's direction. The Cart-Pole has a 4D state and a 1D action space.
% Details
Details on the dynamics of both systems, the reward functions, as well as listings of the domain parameters is given in Appendix~\ref{secapdx_spota_modeling_details}. The nominal models are based on the domain parameter values provided by the manufacturer.

% What we do to use domain randomization
In this paper, both systems have been modeled using the Lagrange formalism and the resulting differential equations are integrated forward in time to simulate the systems.
The associated domain parameters are drawn from a probability distribution $\domparams \sim \domparamdistr{\domparams;\domdistrparams}$, parameterized by $\domdistrparams$ (\eg, mean, variance).
Since randomizing a simulator's physics parameters is not possible right away, so we developed custom a framework to combine \ac{RL} and domain randomization. Essentially, the base environment is modified by wrappers which, \eg, vary the mass or delay the actions.

% -------------------------------------------------- %
\subsection{Experiments Description}
\label{sec_experiments_description}
% -------------------------------------------------- %
The experiments are split into two parts.
At first, we examine the evolution of the \ac{UCBOG} during training (Section~\ref{sec_simtosim_results}). Next, we compare the transferability of the obtained policies across different realizations of the domain, \ie, simulator (Section~\ref{sec_simtosim_results}).
Finally, we evaluate the policies on the real-world platforms (Section~\ref{sec_simtoreal_results}).

\iffalse
\begin{figure*}[t!]
	\centering
	\begin{subfigure}[t]{0.495\textwidth}
		\centering
		\includegraphics[height=4.7cm,keepaspectratio]{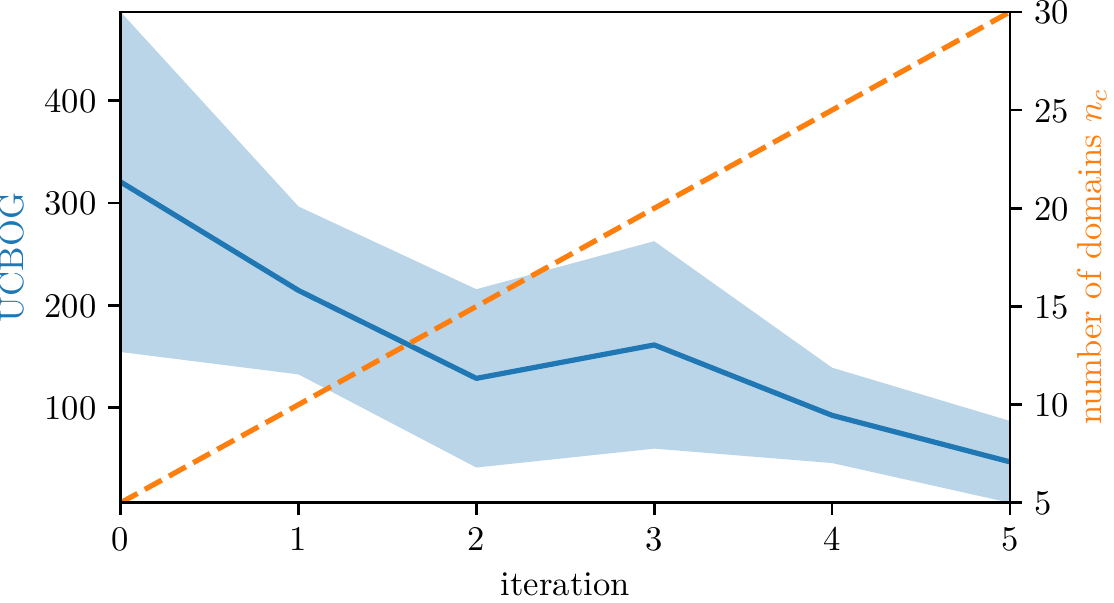}%[height=X.XXcm,keepaspectratio] or [width=\columnwidth]
		\vspace*{-0.5ex} \caption{\vspace*{-0.7ex} Ball-Balancer}
		\label{fig_UCBOG_decrease_QBB}
	\end{subfigure}%
	\hfill
	\begin{subfigure}[t]{0.495\textwidth}
		\centering
		\includegraphics[height=4.7cm,keepaspectratio]{plots/}%[height=X.XXcm,keepaspectratio] or [width=\columnwidth]
		\vspace*{-0.5ex} \caption{\vspace*{-0.7ex} Cart-Pole}
		\label{fig_UCBOG_decrease_QQ}
	\end{subfigure}%
	\caption{%
		\acf{UCBOG} and number of candidate solution domain over the iteration count of \ac{SPOTA}. Every iteration, the number and domains and hence the sample size is increased. The shaded area visualize $\pm 1$ standard deviation across 17 training runs on the simulated Ball-Balancer.}
	\label{fig_UCBOG_decrease}
\end{figure*}
\else
\begin{figure}[t!]
	\centering
	\includegraphics[width=\columnwidth]{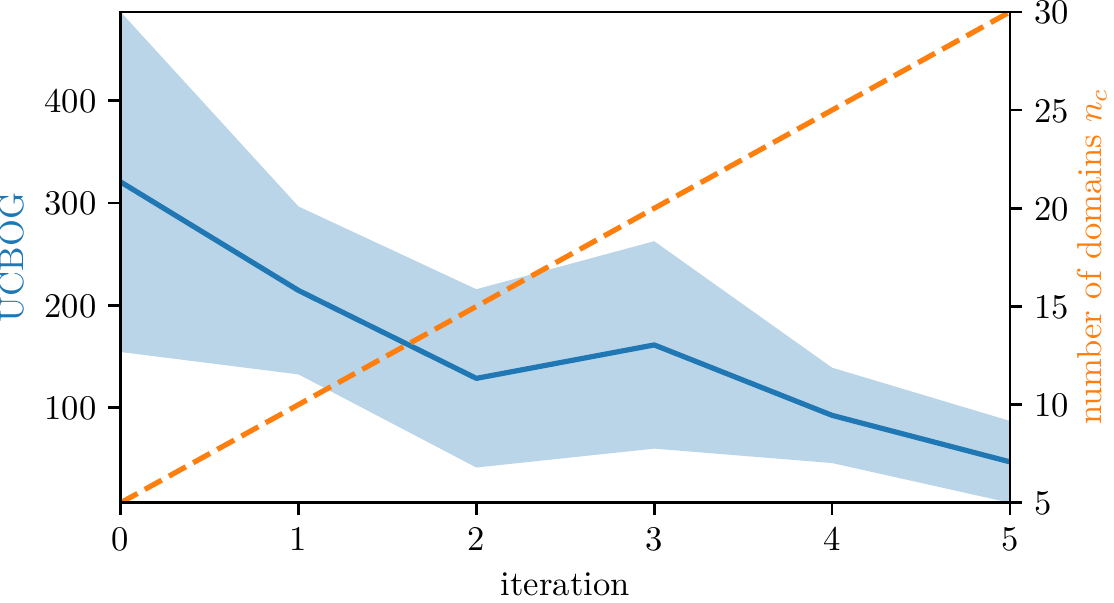}
	\caption{%
		\acf{UCBOG} and number of candidate solution domain over the iteration count of \ac{SPOTA}. Every iteration, the number and domains (dashed line) and hence the sample size is increased. The shaded area visualize $\pm 1$ standard deviation across 9 training runs on the simulated Ball-Balancer.}
	\label{fig_UCBOG_decrease}
\end{figure}
\fi

% Ball-Balancer
For the experiments on the real Ball-Balancer, we choose 8 initial ball positions equidistant from the plate's center and place the ball at these points using a PD controller. As soon as a specified accuracy is reached, the evaluated policy is activated for 2000 time steps, \ie, 4 seconds.
% Cart-Pole
All experiments on the real Carl-pole start with the cart centered on the rail and the pendulum pole hanging down. After calibration, the pendulum is swung up using an energy-based controller. When the system's state is within a specified threshold, evaluated policy is executed for 4000 time steps, \ie, 8 seconds.

% Details about the training ect
All policies %obtained by \ac{RL} algorithms
have been trained in simulation with observation noise to mimic the noisy sensors. To focus on the domain parameters' influence, we executed the \simtosim experiments without observation noise.
The policy update at the core of \ac{SPOTA}, \ac{EPOpt}, and \ac{PPO} is done by the Adam optimizer~\cite{Kingma_etal_2015}.
In the \simtosim experiments, the rewards are computed from the ideal states coming from the simulator, while for the \simtoreal experiments the rewards are calculated from the sensor measurements and their time derivatives.
% Link to the hyper-params
The hyper-parameters chosen for the experiments can be found in the Appendix~\ref{secapdx_hparams}.

% -------------------------------------------------- %
\subsection{Sim-to-Sim Results}
\label{sec_simtosim_results}
% -------------------------------------------------- %

\begin{figure*}[h]
	\centering
	\begin{subfigure}[t]{0.495\textwidth}
		\centering
		\includegraphics[height=5.2cm,keepaspectratio]{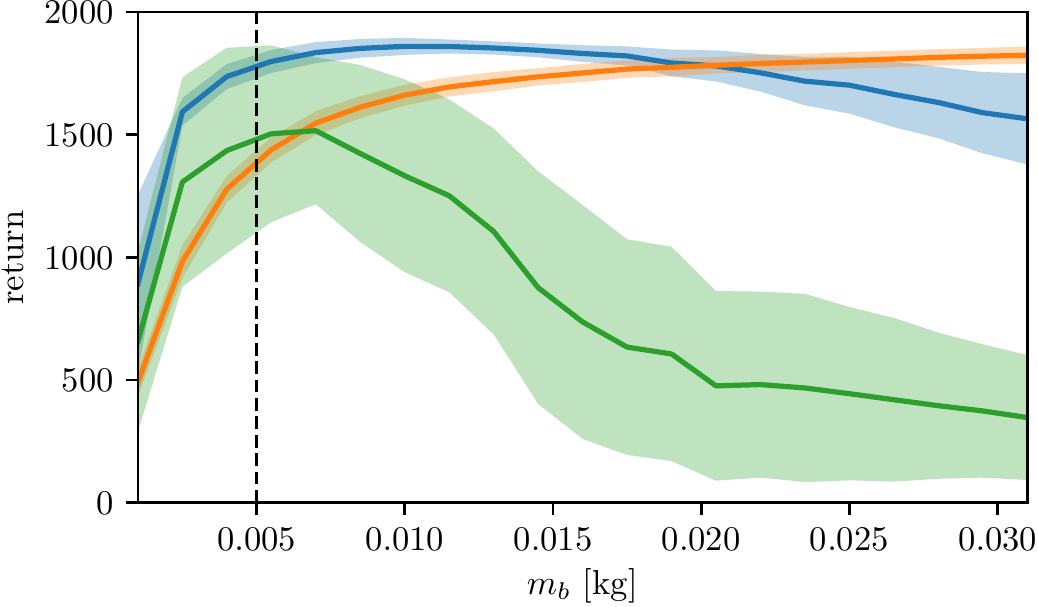}%[height=X.XXcm,keepaspectratio] or [width=\columnwidth]
		\vspace*{-0.5ex} \caption{\vspace*{-0.7ex} Ball Balancer -- varying ball mass}
		\label{fig_QBB_sim_eval_dp_range_m_ball}
	\end{subfigure}%
\hfill
	\begin{subfigure}[t]{0.495\textwidth}
		\centering
		\includegraphics[height=5.2cm,keepaspectratio]{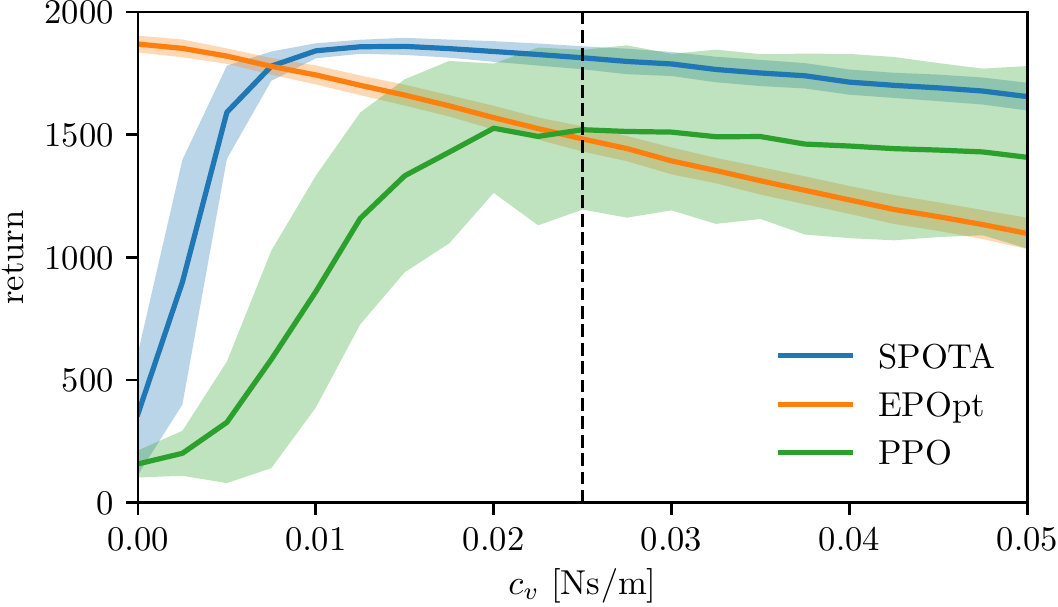}%[height=.XXXcm,keepaspectratio] or [width=\columnwidth]
		\vspace*{-0.5ex} \caption{\vspace*{-0.7ex} Ball Balancer -- varying ball viscous friction coeff.}
		\label{fig_QBB_sim_eval_dp_range_c_frict}
	\end{subfigure}%
\\[1em]
	\begin{subfigure}[t]{0.495\textwidth}
		\centering
		\includegraphics[height=5.2cm,keepaspectratio]{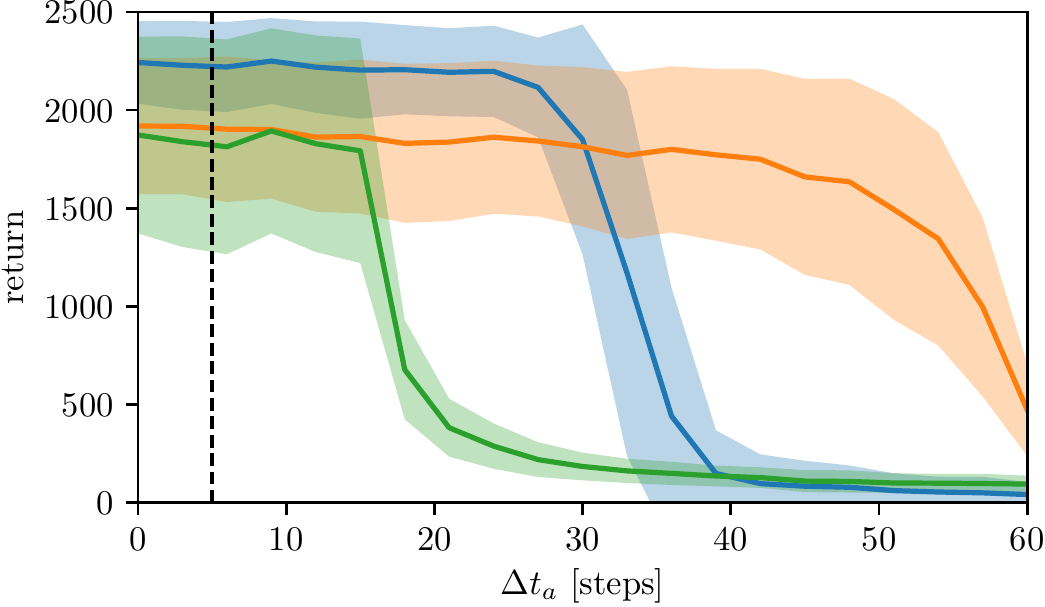}%[height=X.XXcm,keepaspectratio] or [width=\columnwidth]
		\vspace*{-0.5ex} \caption{\vspace*{-0.7ex} Cart-Pole -- varying action delay}
		\label{fig_QCP_sim_eval_dp_range_act_delay}
	\end{subfigure}%
\hfill
	\begin{subfigure}[t]{0.495\textwidth}
		\centering
		\includegraphics[height=5.2cm,keepaspectratio]{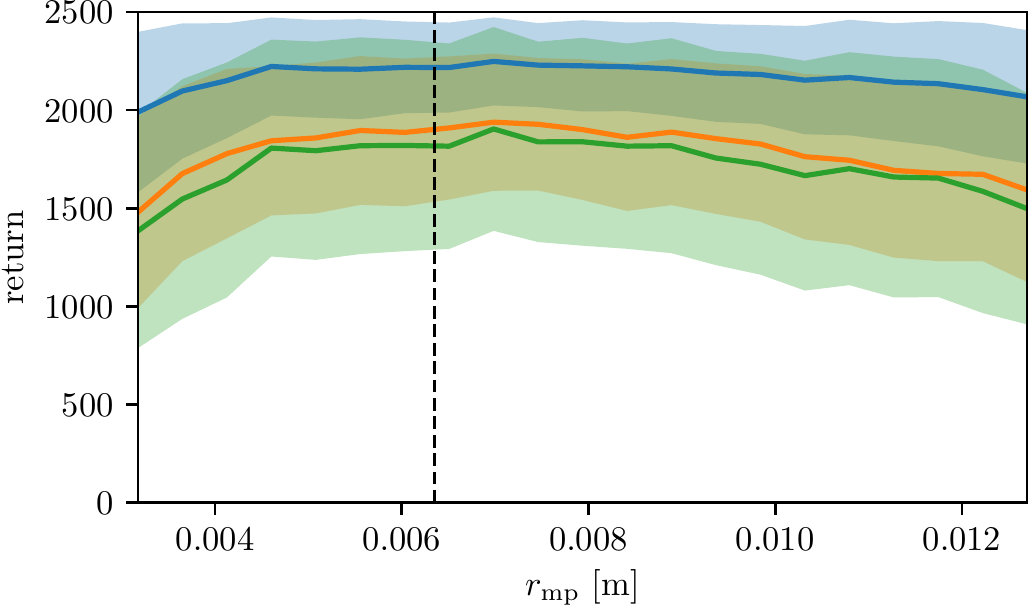}%[height=.XXXcm,keepaspectratio] or [width=\columnwidth]
		\vspace*{-0.5ex} \caption{\vspace*{-0.7ex} Cart-Pole -- varying motor pinoin radius}
		\label{fig_QCP_sim_eval_dp_range_r_mp}
	\end{subfigure}%
	\caption{%
		Evaluation of the learned control policies on the simulated Ball Balancer (top row) as well as the simulated Cart-Pole (bottom row), (a) varying the ball mass $m_b$, (b) the viscous friction coefficient, (c) the action delay $\Delta t_a$, and (d) the motor pinion radius $r_{mp}$.
		Every domain parameter configuration has been evaluated on 360 rollouts with different initial states, synchronized across all policies.
		The dashed lines mark the nominal parameter values. The solid lines represent the means, and shaded areas show $\pm 1$ standard deviation
	}
	\label{fig_QBB_QCP_sim_eval_dp_range}
\end{figure*}

The \ac{UCBOG} value \eqref{eq_UCBOG} at each \ac{SPOTA} iteration depends on several hyper-parameters such as the current number of domains and reference solutions, or the quality of the current iteration's solution candidate.
Figure~\ref{fig_UCBOG_decrease} displays the descent of the \ac{UCBOG} as well as the growth of the number of domains with the increasing iteration count of \ac{SPOTA}.
% Decrease only in expectation
As described in Section~\ref{sec_OG_decrease}, the \ac{OG} and thus the \ac{UCBOG} only decreases in expectation. Therefore, it can happen that for a specific training run the \ac{UCBOG} increases from one iteration to the next.
% Ratio neg OG samples
Moreover, we observed that the proportion of negative \ac{OG} estimates \eqref{eq_optgap_sample} increases with progressing iteration count.
This observation can be explained by the fact that \ac{SPOTA} increases the number of domains used during training. Hence the set's empirical mean approximates the domain parameter distribution $\domparamdistr{\domparams;\domdistrparams}$ progressively better, \ie, the candidate and reference solution become more similar.
% Merged plot
Note, that due to the computational complexity of \ac{SPOTA} we decided to combine results from experiments with different hyper-parameters in Figure~\ref{fig_UCBOG_decrease}.

To estimate the robustness \wrt model parameter uncertainties, we evaluate policies trained by \ac{SPOTA}, \ac{EPOpt}, and \ac{PPO} under on multiple simulator instances, varying only one domain parameter.
The policies' sensitivity to different parameter values is displayed in the Figure~\ref{fig_QBB_QCP_sim_eval_dp_range}.
% Interpretation of the sim-to-sim results
From the Figures~\ref{fig_QBB_sim_eval_dp_range_m_ball}~to~\ref{fig_QCP_sim_eval_dp_range_act_delay} we can see that the policies learned using (plain) \ac{PPO} are less robust to changes in the domain parameter values. In contrast, \ac{SPOTA} and \ac{EPOpt} are able to maintain their level of performance across a wider range of parameter values.
The Figure~\ref{fig_QCP_sim_eval_dp_range_r_mp} shows the case of a domain parameter to which all policies are equally insensitive. 
We can also see that \ac{EPOpt} trades off performance in the nominal domains for performance in more challenging domains (\eg, low friction). This behavior is a consequence of its \mbox{\acs{CVaR}-based} objective function~\cite{Rajeswaran_etal_2017}.
Moreover, the results % 
show a higher variance for the \ac{PPO} policy than for the others. From this, we conclude that domain randomization also acts as regularization mechanism. 
The final \ac{UCBOG} value of the evaluated \ac{SPOTA} policies was \num{46.42} for the Ball-Balancer % 2019-05-07_10-31-44--fs-250_fnn-32-32-tanh_obsnoise-s_actdelay-30_warmstart_w-let
and \num{55.14} for the Cart-Pole.
Note, that the \ac{UCBOG} can not be directly observed from the curves in Figure~\ref{fig_QBB_QCP_sim_eval_dp_range}, since the \ac{UCBOG} reflects the gap in performance between the best policy for a specific simulator instance and the candidate policy, whereas the Figure~\ref{fig_QBB_QCP_sim_eval_dp_range} only shows the candidates' performances.

% -------------------------------------------------- %
\subsection{Sim-to-Real Results}
\label{sec_simtoreal_results}
% -------------------------------------------------- %
\begin{figure*}[t!]
	\centering
	\begin{subfigure}[t]{0.495\textwidth}
		\centering
		\includegraphics[height=4.7cm,keepaspectratio]{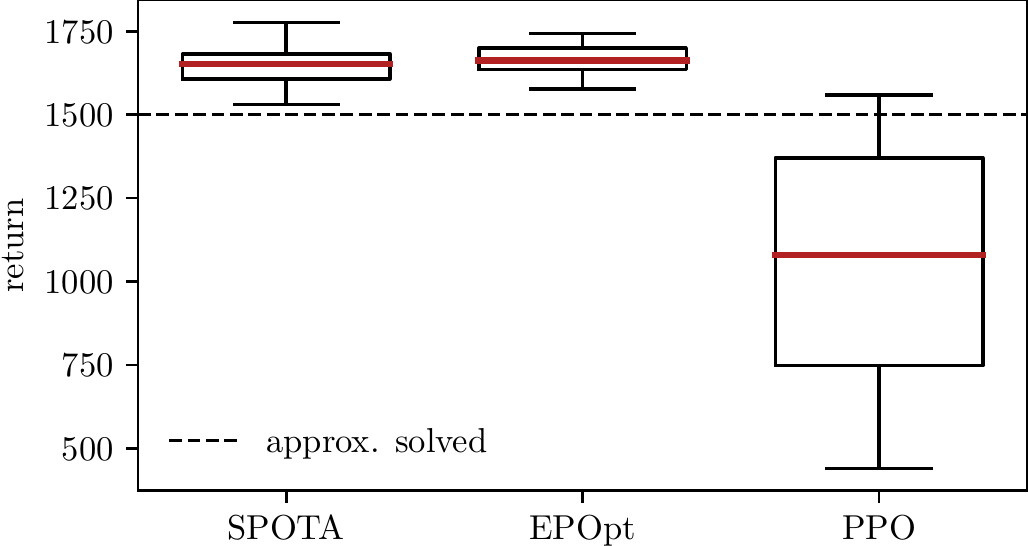}%[height=X.XXcm,keepaspectratio] or [width=\columnwidth]
		\vspace*{-0.5ex} \caption{\vspace*{-0.7ex} Ball Balancer -- \simtoreal}
		\label{fig_real_eval_QBB}
	\end{subfigure}%
\hfill
	\begin{subfigure}[t]{0.495\textwidth}
		\centering
	        \includegraphics[height=4.7cm,keepaspectratio]{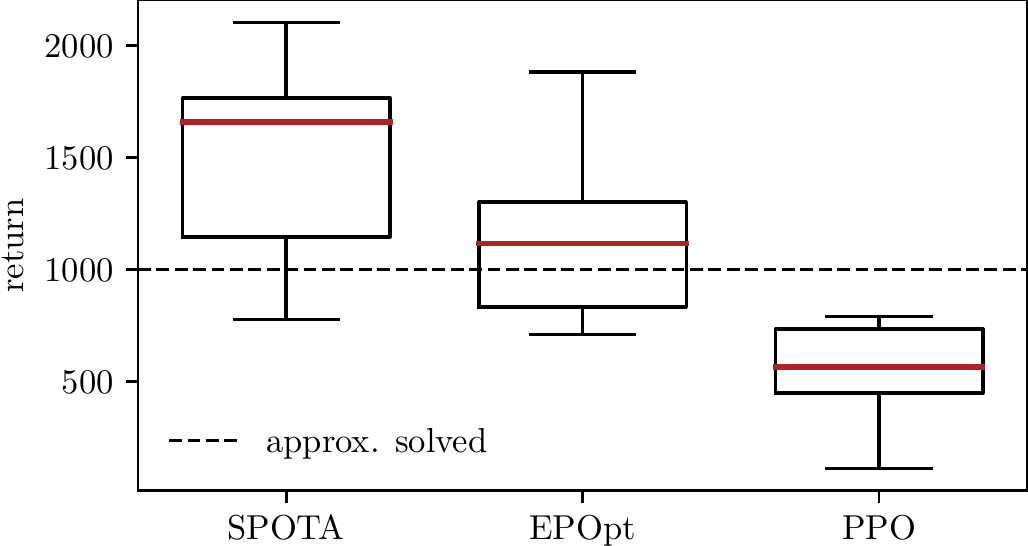}%[height=X.XXcm,keepaspectratio] or [width=\columnwidth]
		\vspace*{-0.5ex} \caption{\vspace*{-0.7ex} Cart-Pole -- \simtoreal}
		\label{fig_real_eval_QCP}
	\end{subfigure}%
	\caption{%
		Evaluation of the learned control policies on (a) the real Ball Balancer and (b) the real Cart-Pole.
		The results were obtained from 40 rollouts per policy on the Ball Balancer (5 repetitions for 8 different initial ball positions) as well as \mbox{10 rollouts} (1 initial state) on the Cart-Pole.
		The dashed lines approximately mark the threshold where the tasks are solved, \ie, the ball is centered in the middle (a), or the pendulum is stabilized upright (b) at the end of the episode.
		%The mean is displayed as a purple diamond.
	}
	\label{fig_real_eval}
\end{figure*}

When transferring the policies from simulation to reality without any fine-tuning, we obtain the results reported in Figure~\ref{fig_real_eval}.
The approaches using domain randomization are in most cases able to directly transfer from simulation to reality.
In contrast, the policies trained on a singular nominal model using failed to transfer in all but 2 trials on the Ball-Balancer as well as in all trials on the Cart-Pole, even though these policies solved the simulated environments.

Regarding the Ball-Balancer, one explanation why the reported \ac{PPO} policy did not transfer to the real platform could be the value of the viscous friction coefficient (Figure~\ref{fig_QBB_sim_eval_dp_range_c_frict}).
A possible test for this hypothesis would be to train multiple policies on a range of nominal models with altered viscous friction parameter value, and if these policies still do not transfer, examine the next domain parameter. However, this procedure is redundant and can quickly become prohibitively time-intensive.
%We believe that the reported \simtoreal results show that learning from randomized simulations is a promising alternative.
%
Concerning the experiments on the Cart-Pole, we observe a larger reality-gap for all policies.
We believe that this discrepancy is caused by unmodeled effects between the rack and the cart's pinion (\eg, the heavy wear and tear of the pinon made out of plastic).
%In particular, \ac{EPOpt}
Moreover, the variance of the returns is significantly higher. This increase can be largely explained by the variance in the initial states caused by the pre-defined swing-up controller.

% Refereing to the \ac{UCBOG} values reported in Section~\ref{sec_simtosim_results}, we follow that the qulity of the tarnsferability assessments differs natably.
% While the \ac{UCBOG} for the Ball-Balancer selected policy $\optgapmean[\refsym][U]{\polparamsoptcand}}$ equals appro
% the simulations' inability to cover the real system dynamics.

A video of the \ac{SPOTA} policy's \simtoreal transfer on both platforms can be found at \url{https://www.ias.informatik.tu-darmstadt.de/Team/FabioMuratore}.%https://www.youtube.com/channel/UC7YoKtcqHYq3wx9K1MlbpRw

% -------------------------------------------------- %
\subsection{Limitations of the Presented Method}
\label{sec_limitations}
% -------------------------------------------------- %
The computation of the \ac{UCBOG} \eqref{eq_UCBOG}, and hence the estimation of the \ac{SOB}, relies on the relative quality of the candidate and the reference solutions. Therefore, the most notable limitation of the presented method is the optimizer's ability to reliably solve the \ac{SP} \eqref{eq_SPn}.
Since we are interested in the general setting of learning a control policy from a black-box simulator, we chose a model-free \ac{RL} algorithm. These kind of algorithms can not guarantee to find the globally optimal, or loosely speaking a very good, solution.
One way to alleviate this problem is to compute the reference policies from a single domain using methods from control engineering, \eg a \acs{LQR}.
However, this solution would require an analytic model of the system and a specific type of reward function to preserve comparability between the solutions, \eg quadratic functions in case of the \acs{LQR}.

Another limitation of \ac{SPOTA} is the increased hyper-parameter space which is a direct consequence from the employed (static) domain randomization. In combination with the fact that \ac{SPOTA} is solving the underlying \ac{RL} task $(1 + n_G)n_{\text{iter}}$ times, the procedure becomes computationally expensive.
One can counter this problem by parallelizing the computation of the reference policies as well as the hyper-parameter search. Both are embarrassingly parallel.

Moreover, \ac{SPOTA} does not considers uncertainty in the parametrization of the domain parameter distribution $\domparamdistr{\domparams; \domdistrparams}$.
One possibility to tackle this potential deficiency is to adapt these distributions, as for example done in \cite{Chebotar_etal_2019}.
Moving from parametric to nonparametric models of the domain parameter distribution is easily possible since \ac{SPOTA} only requires to sample from them.

Finally, to estimate the \ac{SOB}, \ac{SPOTA} assumes that the target domain is covered by the source domain distribution, which can not be guaranteed if the target is a real-world system.
However, in the current \sota there is no way to estimate a policy's transferability to a domain from an unknown distribution.
Due to mentioned assumption, \ac{SPOTA}'s transferability assessment strongly depends on the simulator's ability to model the real world. %This dependency is reflected by the mismatch between the low final \ac{UCBOG} value for the Cart-Pole's \ac{SPOTA} policy and the policy's real-world performance.

% %%%%%%%%%%%%%%%%%%%%%%%%%%%%%%%%%%%%%%%%%%%%%%%%%% %
\section{Related Work}
\label{sec_related_work}
% %%%%%%%%%%%%%%%%%%%%%%%%%%%%%%%%%%%%%%%%%%%%%%%%%% %
In the following, we review excerpts of the literature regarding the transfer of control policies from simulation to reality, the concept of the optimality gap in \acfp{SP}, and the application of randomized physics simulations.
This paper is a substantial extension of our previous work~\cite{Muratore_etal_2018}, adding the derivation of the \ac{SOB} from the \ac{OG} (Section~\ref{sec_SOB}~to~\ref{sec_connection_OG_SOB}), an outlier rejection component (Algorithm~\ref{algo_SPOTA}), and the method's first real-world verification using two experiments (Section~\ref{sec_simtoreal_results}).

% -------------------------------------------------- %
\subsection{Key publications on the Optimality Gap}
% -------------------------------------------------- %
% Samle based optimization is optimistically biased
Hobbs~and~Hepenstal~\cite{Hobbs_Hepenstal_1989} proved for linear programs that optimization is optimistically biased, given that there are errors in estimating the objective function coefficients. Furthermore, they demonstrated the \enquote{optimistic bias} of a nonlinear program, and mentioned the effect of errors on the parameters of linear constraints.
% sPOTA fulfills the assumptions
The optimization problem introduced in Section~\ref{sec_SPOTA} belongs to the class of \acp{SP} for which the assumption required in \cite{Hobbs_Hepenstal_1989} are guaranteed to hold.
% sAA methods
The most common approaches to solve convex \acp{SP} are sample average approximation methods, including:
(i) the \acl{MRP} and its derivatives~\cite{Mak_etal_1999,Bayraksan_Morton_2006} which assess a solution's quality by comparing with sampled alternative solutions, and
(ii) \acl{RA}~\cite{Pasupathy_Schmeiser_2009,Kim_etal_2015} which iteratively improved the solution by lowering the error tolerance.
Bastin~\etal~\cite{Bastin_etal_2006} extended the existing convergence guarantees from convex to non-convex \acp{SP}, showing almost sure convergence of the minimizers. % under mild assumptions.

% -------------------------------------------------- %
\subsection{Prior work on the Reality Gap} 
% -------------------------------------------------- %
% Physics simulation and the resulting reality gap as the resulting problem
Physics simulations have already been used successfully in robot learning. Traditionally, simulators are operating on a single nominal model, which makes the direct transfer of policies from simulation to reality highly vulnerable to model uncertainties and biases. Thus, model-based control in most cases relies on fine-tuned dynamics models.
\\
% Viewpoints on the sim-real missmatch
The mismatch between the simulated and the real world has been addressed by robotics researchers from different viewpoints. Prominent examples are:
\begin{enumerate}
	\item adding \iid noise to the observations and actions in order to mimic real-world sensor and actuator behavior~\cite{Jakobi_etal_1995},%omitted Jakobi_1997
	\item repeating model generation and selection depending on the short-term state-action history~\cite{Bongard_etal_2006},
	\item learning a transferability function which maps solutions to a score that quantifies how well the simulation matches the reality~\cite{Koos_etal_2013},
	\item adapting the simulator to better match the observed real-world data~\cite{Chebotar_etal_2019,Hanna_Stone_2017},
	\item randomizing the physics simulation's parameters, and
	\item applying adversarial perturbations to the system,
\end{enumerate}
where the last two approaches are particularly related and discussed in the  Sections~\ref{sec_related_work_dr}~and~\ref{sec_related_work_ap}.
The fourth point comprises methods based on system identification, which conceptually differ from the presented method since these seek to find the simulation closest to reality, \eg minimal prediction error.
A recent example in the field of robotics is the work by Hanna and~Stone~\cite{Hanna_Stone_2017}, where an action transformation is learned such that the transformed actions applied in simulation have the same effects as applying the original actions had on the real system.

% -------------------------------------------------- %
\subsection{Required Randomized Simulators}
\label{sec_related_work_simulators}
% -------------------------------------------------- %
Simulators can be obtained by implementing a set of physical laws or by using general purpose physics engines.
% as well as by using general purpose physics engines like Bullet~\cite{Bullet_physics_engine}, MuJoCo~\cite{Mujoco_physics_engine}, or Vortex~\cite{Vortex_physics_engine}.
The associated physics parameters can be estimated by system identification, which involves executing control policies on the physical platform~\cite{Isermann_Munchhof_2010}. 
% Gauss-Markov theorem
Additionally, using the Gauss-Markov theorem one could also compute the parameters' covariance and hence construct a normal distribution for each domain parameter. % Cov(\domparams) = (X\tran X)\inv * sigma²
% Use GPs for learning the dyamics
Alternatively, the system dynamics can be captured using nonparametric methods like Gaussian processes \cite{Rasmussen_Williams_2006}.
%Another alternative could be to come up with a method that takes a camera stream as input, segments the objects in the scene, and returns a probability distribution for each domain parameter. But to the best of the authors' knowledge, no such method has been presented, yet.
% Always flawed
It is important to keep in mind, that even if the selected procedure yields a very accurate model parameter estimate, simulators are nevertheless just approximations of the real world and are thus always flawed.

% Domain param distr as physically plausible prior
As done in~\cite{Mordatch_etal_2015,Rajeswaran_etal_2017,Pinto_etal_2017b,Peng_etal_2018,Yu_etal_2017} we use the domain parameter distribution as a prior which ensures the physical plausibility of each parameter.
Note that specifying this distribution in the current state-of-the-art requires the researcher to make design decisions. 
% Static vs. adaptive distr
Chebotar~\etal~\cite{Chebotar_etal_2019} presented a promising method which adapts the domain parameter distribution using real-world data in the loop.
The main advantage is that this approach alleviates the need for hand-tuning the distributions of the domain parameters, which is currently a significant part of the hyper-parameter search. However, the initial distribution still demands for design decisions.
On the downside, the adaptation requires data from the real robot which is considered significantly more expensive to obtain.
Since we aim for performing a \simtoreal transfer without using any real-world data, the introduced method only samples from static probability distributions.

% -------------------------------------------------- %
\subsection{Background on Domain Randomization}
\label{sec_related_work_dr}
% -------------------------------------------------- %
% Domain randomization as a solution to bridge the rality gap
There is a large consensus that further increasing the simulator's accuracy alone will not bridge the reality gap.
Instead, the idea of domain randomization has recently gained momentum. The common characteristic of such approaches is the perturbation of the parameters which determine the physics simulator and the state estimation, including but not limited to the system dynamics.
% Domain randomization in (robot) RL
While the idea of randomizing the sensors and actuators dates back to at least 1995~\cite{Jakobi_etal_1995}, the systematic analysis of perturbed simulations in robot \ac{RL} is a relatively new research direction. 
%~\cite{Wang_etal_2010,Antonova_Cruciani_2017,Mordatch_etal_2015,Rajeswaran_etal_2017,Peng_etal_2018,Tobin_etal_2017a,Pinto_etal_2017b,Mandlekar_etal_2017,Pinto_etal_2017a,Kurutach_etal_2018,Sadeghi_Levine_2017,Bousmalis_etal_2018,Matas_etal_2018}.

Wang,~Fleet,~and~Hertzmann~\cite{Wang_etal_2010} proposed sampling initial states, external disturbances, goals, as well as actuator noise from probability distributions and learned walking policies in simulation.
Regarding robot \ac{RL}, recent domain randomization methods focus on perturbing the parameters defining the system dynamics. Approaches cover:
(i) trajectory optimization on finite model-ensembles~\cite{Mordatch_etal_2015}
(ii) learning a feedforward \acs{NN} policy for an under-actuated problem~\cite{Antonova_Cruciani_2017},
(iii) using a risk-averse objective function~\cite{Rajeswaran_etal_2017},
(iv) employing recurrent \acs{NN} policies trained with experience replay~\cite{Peng_etal_2018}, and
(v) optimizing a policy from samples of a model randomly chosen from a set which is repeatedly fitted to real-world data~\cite{Kurutach_etal_2018}.
From the listed approaches \cite{Mordatch_etal_2015,Antonova_Cruciani_2017,Peng_etal_2018} were able to cross the reality gap without acquiring samples from the real world.

% Domain randomization in computer vision
Moreover, there is a significant amount of work applying domain randomization to computer vision.
One example is the work by Tobin~\etal~\cite{Tobin_etal_2017a} where an object detector for robot grasping is trained using multiple variants of the environment and applied to the real world.
The approach presented by Pinto~\etal~\cite{Pinto_etal_2017b} combines the concepts of randomized environments and actor-critic training, enabling the direct \simtoreal transfer of the abilities to pick, push, or move objects.
Sadeghi~and~Levine~\cite{Sadeghi_Levine_2017} achieved the \simtoreal transfer by learning to fly a drone in visually randomized environments. The resulting deep \acs{NN} policy was able to map from monocular images to normalized 3D drone velocities.
In~\cite{Matas_etal_2018}, a deep \acs{NN} was trained to manipulate tissue using randomized vision data and the full state information.
% Domain Adaptaion
By combing generative adversarial networks and domain randomization, Bousmalis~\etal~\cite{Bousmalis_etal_2018} greatly reduced the number of necessary real-world samples for learning a robotic grasping task.

% Relation to multi-task learning
Domain randomization is also related to multi-task learning in the sense that one can view every instance of the randomized source domain as a separate task.
In contrast to multi-task learning approaches as presented in~\cite{Deisenroth_etal_2014,Andrychowicz_etal_2017}, a policy learned with \ac{SPOTA} does not condition on the task. Thus, during execution there is no need to infer the task \ie domain parameters.

% -------------------------------------------------- %
\subsection{Randomization Trough Adversarial Perturbations} % Adversarial Perturbations for Domain Randomization
\label{sec_related_work_ap}
% -------------------------------------------------- %
Another method for learning robust policies in simulation is to apply adversarial disturbances to the training process.
Mandlekar~\etal~\cite{Mandlekar_etal_2017} proposed physically plausible perturbations by randomly deciding when to add a rescaled gradient of the expected return.
Pinto~\etal~\cite{Pinto_etal_2017a} introduced the idea of a second agent whose goal is to hinder the first agent from fulfilling its task. Both agents are trained simultaneously and make up a zero-sum game.
In general, adversarial approaches may provide a particularly robust policy. However, without any further restrictions, it is always possible create scenarios in which the protagonist agent can never win, \ie, the policy will not learn the task.

% %%%%%%%%%%%%%%%%%%%%%%%%%%%%%%%%%%%%%%%%%%%%%%%%%% %
\section{Conclusion}
\label{sec_conclusion}
% %%%%%%%%%%%%%%%%%%%%%%%%%%%%%%%%%%%%%%%%%%%%%%%%%% %

% -------------------------------------------------- %
%\subsection{Summary}
%\label{sec_summary}
% -------------------------------------------------- %
% What we added
We proposed a novel measure of the \acf{SOB} for quantifying the transferability of an arbitrary policy learned from a randomized source domain to an unknown target domain from the same domain parameter distribution.
Based on this measure of the \ac{SOB}, we developed a policy search meta-algorithm called \acf{SPOTA}.
This gist of \ac{SPOTA} is to iteratively increase the number of domains and thereby the sample size per iteration until an approximate probabilistic guarantee on the optimality gap holds.
The required approximation of the optimality gap is obtained by comparing the current candidate policy against multiple reference policies evaluated in the associated reference domains.
After training, we can make an approximation of the resulting policy's suboptimality when transferring to a different domain from the same (source) distribution.
% What we could show
To verify our approach we conducted two \simtoreal experiments on second order nonlinear continuous control systems.
The results showed that \ac{SPOTA} policies were able to directly transfer from simulation to reality while the baseline without domain randomization failed.

% -------------------------------------------------- %
%\subsection{Future Work}
%\label{sec_future_work}
% -------------------------------------------------- %
In the future we will investigate different strategies for sampling the domain parameters to replace the \iid sampling from hand-crafted distributions. One idea is to employ Bayesian optimization for selecting the next set of domain parameters.
Thus, the domain randomization could be executed according to an objective, and potentially increase sample-efficiency.
Furthermore, we plan to devise a formulation which frames domain randomization and policy search in one optimization problem. This would allow for an joint treatment of finding a policy and matching the simulator to the real world.

% %%%%%%%%%%%%%%%%%%%%%%%%%%%%%%%%%%%%%%%%%%%%%%%%%% %
\section*{Acknowledgments}
% %%%%%%%%%%%%%%%%%%%%%%%%%%%%%%%%%%%%%%%%%%%%%%%%%% %
Fabio Muratore gratefully acknowledges the financial support from \acl{HRIE}.
\\
Jan Peters received funding from the European Union’s Horizon 2020 research and innovation programme under grant agreement No 640554.

% %%%%%%%%%%%%%%%%%%%%%%%%%%%%%%%%%%%%%%%%%%%%%%%%%% %
% References
% %%%%%%%%%%%%%%%%%%%%%%%%%%%%%%%%%%%%%%%%%%%%%%%%%% %
\iffalse
	\printbibliography
\else
	\bibliographystyle{IEEEtran}
	\bibliography{IEEEabrv,fMRT_arxiv.bib}
\fi

% %%%%%%%%%%%%%%%%%%%%%%%%%%%%%%%%%%%%%%%%%%%%%%%%%% %
% Biographies
% %%%%%%%%%%%%%%%%%%%%%%%%%%%%%%%%%%%%%%%%%%%%%%%%%% %
\iffalse
	\begin{IEEEbiography}[%
		{\includegraphics[width=1in,height=1.25in,clip,keepaspectratio]{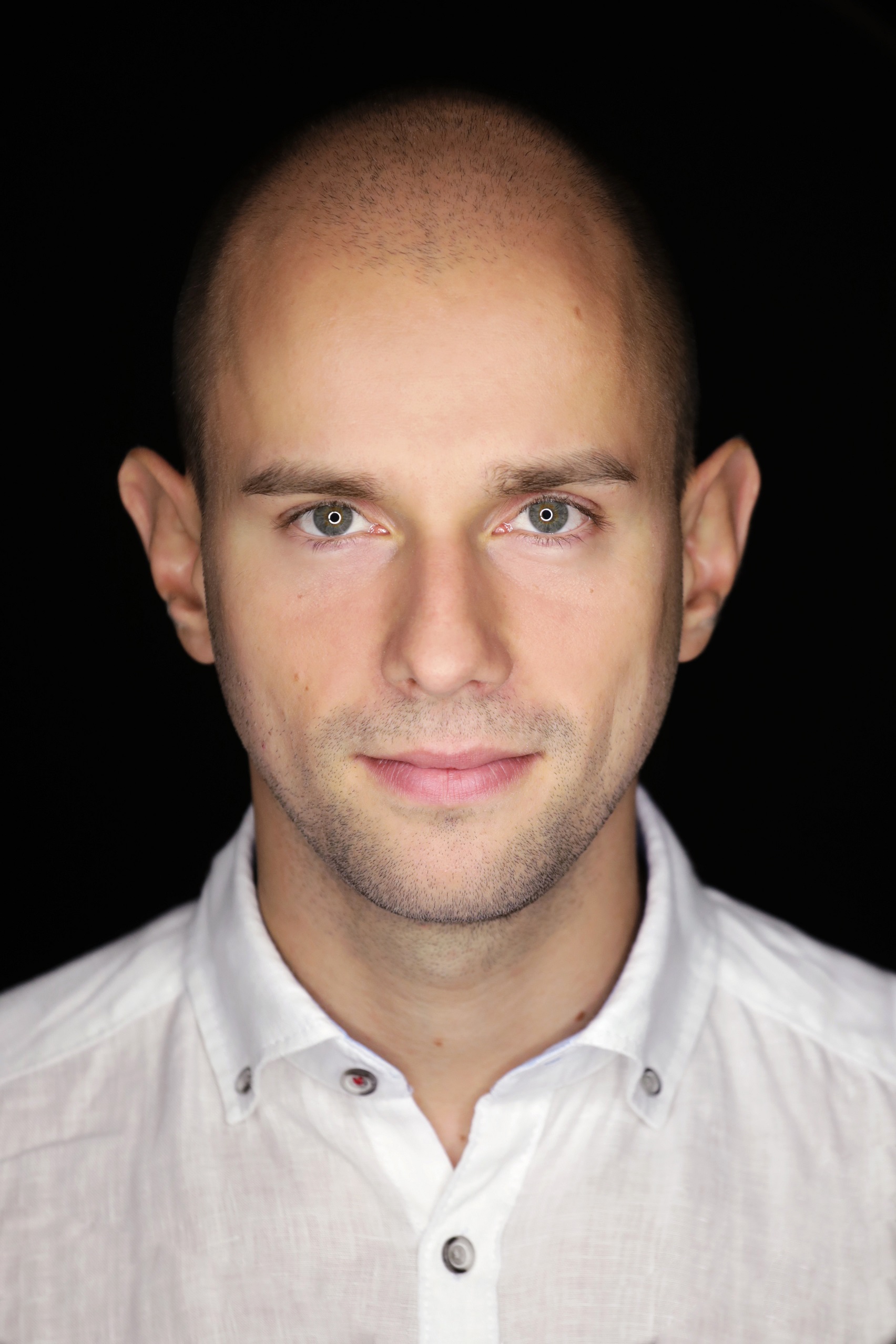}}]
		{Fabio Muratore}
	received his M.Sc. degree in Mechanical Engineering from the Technical University of Munich in 2016.
	Currently, he is perusing his Ph.D. at the Intelligent Autonomous Systems Group at the Computer Science Department of the Technical University of Darmstadt.
	His research interests center around robotics, artificial intelligence, and learning from physics simulations.
	\end{IEEEbiography}
	\vspace*{-1em}
	\begin{IEEEbiography}[%
		{\includegraphics[width=1in,height=1.25in,clip,keepaspectratio]{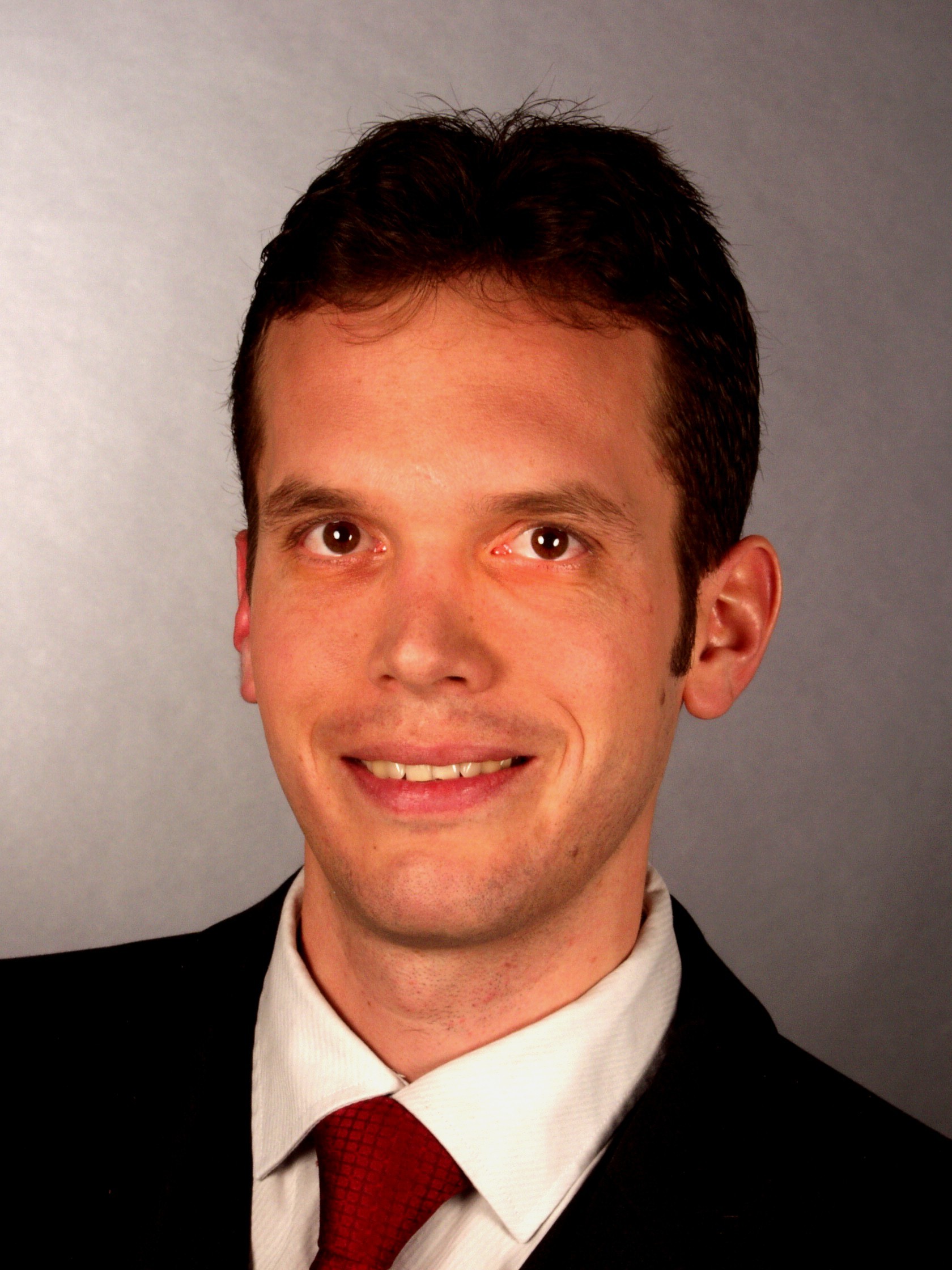}}]
		{Michael Gienger}
	received the diploma degree in Mechanical Engineering from the Technical University of Munich (TUM) in 1998.
	From 1998 to 2003, he was research assistant at the Institute of Applied Mechanics of the TUM, addressing issues in design and realization of biped robots.
	He received his Ph.D. degree with a dissertation on ``Design and Realization of a Biped Walking Robot''.
	After this, Michael Gienger joined the Honda Research Institute Europe in 2003.
	Currently, he works as a Chief Scientist in the field of robotics.
	His research interests include mechatronics, robotics, control systems, robot learning and human-robot interaction. 
	\end{IEEEbiography}
	\vspace*{-1em}
	\begin{IEEEbiography}[%
		{\includegraphics[width=1in,height=1.0in,clip,keepaspectratio]{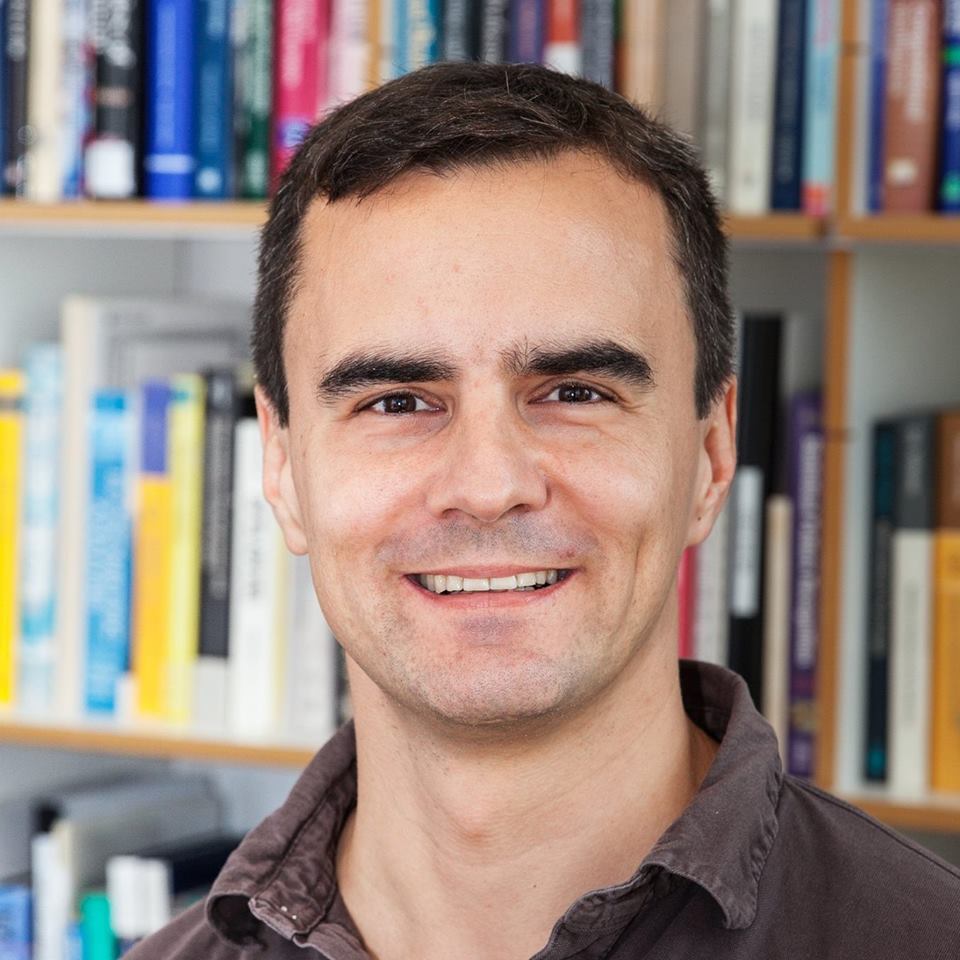}}]
		{Jan Peters}
	is a full professor (W3) for Intelligent Autonomous Systems at the Computer Science Department of the Technical University of Darmstadt, and at the same time a senior research scientist and group leader at the Max-Planck Institute for Intelligent Systems, where he heads the interdepartmental Robot Learning Group.
	Jan Peters has received the Dick Volz Best 2007 US Ph.D. Thesis Runner-Up Award, the Robotics: Science \& Systems - Early Career Spotlight, the INNS Young Investigator Award, and the IEEE Robotics \& Automation Society's Early Career Award as well as numerous best paper awards.
	In 2015, he received an ERC Starting Grant and in 2019, he was appointed as an IEEE Fellow.
	%
	%Jan Peters has studied Computer Science, Electrical, Mechanical and Control Engineering at the Technical University of Munich and FernUni Hagen in Germany, at the National University of Singapore and the University of Southern California (USC). He has received four Master's degrees in these disciplines as well as a Computer Science PhD from USC.
	%Jan Peters has performed research in Germany at DLR, TU Munich and the Max Planck Institute for Biological Cybernetics (in addition to the institutions above), in Japan at the Advanced Telecommunication Research Center (ATR), at USC and at both NUS and Siemens Advanced Engineering in Singapore. 
	\end{IEEEbiography}
\else
	\newpage
\fi

% %%%%%%%%%%%%%%%%%%%%%%%%%%%%%%%%%%%%%%%%%%%%%%%%%% %
% Appendices
\appendices
% %%%%%%%%%%%%%%%%%%%%%%%%%%%%%%%%%%%%%%%%%%%%%%%%%% %

% %%%%%%%%%%%%%%%%%%%%%%%%%%%%%%%%%%%%%%%%%%%%%%%%%% %
\section{Modeling Details on the Platforms}
\label{secapdx_spota_modeling_details}
% %%%%%%%%%%%%%%%%%%%%%%%%%%%%%%%%%%%%%%%%%%%%%%%%%% %
The Ball-Balancer is modeled as a nonlinear second-order dynamical system
\begin{equation}
\label{eq_EoM_QBB}
\ddot{\fs} =
\begin{bmatrix}
\ddot{\theta}_x\\ 
\ddot{\theta}_y\\ 
\ddot{x}_b\\ 
\ddot{y}_b
\end{bmatrix} = 
\begin{bmatrix}
(A_m V_x - B_v \dot{\theta}_x) / J_{eq}\\ 
(A_m V_y - B_v \dot{\theta}_y) / J_{eq}\\ 
(-c_v \dot{x}_b r_b^2 - J_b r_b \ddot{\alpha} +  m_b x_b \dot{\alpha}^2 r_b^2 \\ \quad +~c_{kin} m_b g r_b^2 \sin{\theta_x} ) / \zeta\\ 
(-c_v \dot{y}_b r_b^2 - J_b r_b \ddot{\beta}  +  m_b y_b \dot{\beta}^2 r_b^2 \\ \quad +~c_{kin} m_b g r_b^2 \sin{\theta_y} ) / \zeta
\end{bmatrix}, 
\end{equation}
with the motor shaft angles $\theta_x$ and $\theta_y$, the ball positions relative to the place center $x_b$ and $y_b$, the plate angle $\beta$ and $\alpha$ around the $x$ and $y$ axis, and the commanded motor voltages ${\fa\tran = \left[V_x, V_y\right]}$.
To model the gears' backlash, we set all voltage values between $V_{thold,-}$ and $V_{thold,+}$ to zero. These threshold values have been determined in separate experiments for both servo motors.
The Ball-Balancer's domain parameters as well as the ones derived from them are listed in Table~\ref{tab_dom_param_QBB}.
For the Ball-Balancer's we define the reward function as 
\begin{gather}
\label{eq_rewfcn_QBB}
\rewfcn{\fs_t, \fa_t} = \exp{c \left( \fs_t\tran \fQ_{BB} \fs_t + \fa_t\tran \fR_{BB} \fa_t \right) }\\
\text{with }
c = \frac{\ln{r_{min}}}{\max_{\fs\in\stateset[\domparamsnom], \fa\in\actionset[\domparamsnom]} \fs\tran \fQ_{BB} \fs + \fa\tran \fR_{BB} \fa}.
\end{gather}
Given a lower bound for the reward $r_{min} \in [0,1]$, the reward function above yields values within $[r_{min},1]$ at each time step. We found that the scaling constant $c < 0$ is beneficial for the learning procedure, since it prohibits the reward from going to zero too quickly. The constant's denominator can be easily inferred from the nominal state and action set's boundaries.

\iftrue
The Cart-Pole is modeled as a nonlinear second-order dynamical system given by the solution of
\begin{gather}
\label{eq_EoM_QCP}
% A
\begin{bmatrix}
m_p + J_{eq} & m_p l_p \cos{\alpha} \\
m_p l_p \cos{\alpha} & J_p + m_p l_p^2
\end{bmatrix}
% x_ddot
\begin{bmatrix}
\ddot{x}\\ 
\ddot{\alpha}
\end{bmatrix} =\\
% b 
\begin{bmatrix}
F -m_p l_p \sin{\alpha} \alphad^2 - B_{eq} \dot{x}\\
-m_p l_p g \sin{\alpha} - B_p \dot{\alpha}\\
\end{bmatrix}, 
\end{gather}
where the commanded motor voltage $V$ is encapsulated in 
\begin{equation}
F = \frac{\eta_g K_g k_m}{R_m r_{mp}} \left( \frac{\eta_m V - K_g k_m \dot{x}}{r_{mp}} \right).
\end{equation}
The system's state $\fs$ is given by the cart's position $x$ and the pole's angle $\alpha$, which are defined to be zero at the rail's center and hanging down vertically, respectively.
The Cart-Pole's domain parameters as well as the parameters derived from them are listed in Table~\ref{tab_dom_param_QCP}.
\iffalse
% QCP swing-up
For the Cart-Pole we define the reward function as
\begin{equation}
\label{eq_rewfcn_QCP}
\rewfcn{\fs_t} = -c_1 \left(e_\alpha^2 +  \log{e_\alpha^2 + c_2} \right) - c_3 |x| + c_4
\end{equation}
with the modulated pole angle error ${e_\alpha = (\pi - \alpha)~\textrm{mod}~2\pi}$, and the constants $c_1$, $c_2$ and $c_3$. Thus, the reward is in range $]-\infty, -c_1\log{c_2} + c_4]$ for every time step.
\else
% QCP balancing
Similar to the Ball-Balancer, the Cart-Pole's reward function is based on an exponentiated quadratic cost
\begin{equation}
\label{eq_rewfcn_QCP}
\rewfcn{\fs_t, a_t} = \exp{- \left( \fs_t\tran \fQ_{CP} \fs_t + a_t\tran R_{CP} a_t \right) }.
\end{equation}
Thus, the reward is in range $]0, 1]$ for every time step.
\fi

\else
The Qube is modeled as a nonlinear second-order dynamical system given by the solution of
\begin{gather}
\label{eq_EoM_QQ}
% A
\begin{bmatrix}
J_r + m_p l_r^2 + \onefourth m_p l_p^2 (\cos{\alpha})^2 & \onehalf m_p l_p l_r \cos{\alpha} \\
\onehalf m_p l_p l_r \cos{\alpha} & J_p + \onefourth m_p l_p^2
\end{bmatrix}
% x_ddot
\begin{bmatrix}
\ddot{\theta}\\ 
\ddot{\alpha}
\end{bmatrix} =\\
% b 
\begin{bmatrix}
\tau - \onehalf m_p l_p^2 \sin{\alpha} \cos{\alpha} \dot{\theta} \dot{\alpha} - \onehalf m_p l_p l_r \sin{\alpha} \dot{\alpha}^2 - d_r \dot{\theta}\\
-\onefourth m_p l_p^2 \sin{\alpha} \cos{\alpha} \dot{\theta}^2 - \onehalf m_p l_p g \sin{\alpha} - d_p \dot{\alpha}\\
\end{bmatrix}, 
\end{gather}
with the system's state $\fs$ given by the rotary angle $\theta$ and the pendulum angle $\alpha$, which are defined to be zero when the rotary pole is centered and the pendulum pole is hanging down vertically.
The commanded motor voltage $a = V$ controls the servo motor's torque
\begin{equation}
\tau = \frac{k_m \left( V - k_m \dot{\theta} \right)}{R_m}.
\end{equation}
The Qube's domain parameters as well as the parameters derived from them are listed in Table~\ref{tab_dom_param_QQ}.
Similar to the Ball-Balancer, the Qube's reward function is based on an exponentiated quadratic cost
\begin{equation}
\label{eq_rewfcn_QQ}
\rewfcn{\fs_t, a_t} = \exp{- \left( \fs_t\tran \fQ_{Q} \fs_t + a_t\tran R_{Q} a_t \right) }.
\end{equation}
Thus, the reward is in range $]0, 1]$ for every time step.
\fi

\newpage
% %%%%%%%%%%%%%%%%%%%%%%%%%%%%%%%%%%%%%%%%%%%%%%%%%% %
\section{Parameter values for the experiments}
\label{secapdx_hparams}
% %%%%%%%%%%%%%%%%%%%%%%%%%%%%%%%%%%%%%%%%%%%%%%%%%% %
\begin{table}[h]
	\centering
	\caption{Domain parameter values for the illustrative example. Additional (domain-independent) parameters are $m = \SI{1}{\kilogram}$ and $\domdistrparam = 0.7$.}
	\label{tab_params_ill_exmpl}
	\begin{tabular}{llllll}
    \rowcolor{gray!50} \textbf{Domain} & $g_i [\si{\meter\per\sec\squared}]$ & $k_i [\si{\newton/\meter}]$ & $x_i [\si{\meter}]$\\
	Mars                               & \num{3.71}                          & \num{1000}                  & \num{0.5}\\
	Venus                              & \num{8.87}                          & \num{3000}                  & \num{1.5}
\end{tabular}
% Old version
%\newcolumntype{G}{>{\columncolor{gray!50}}l}
%\newcolumntype{g}{>{\columncolor{gray!25}}l}
%	\begin{tabular}{Glglgl}
%	Domain & $g_i [\si{\meter\per\sec\squared}]$ & $k_i [\si{\newton/\meter}]$ & $x_i [\si{\meter}]$\\
%	Mars   & 3.71                                & 1000                        & 0.5                \\
%	Venus  & 8.87                                & 3000                        & 1.5   
%	\end{tabular}
\end{table}
\begin{table}[hb]
	\centering
	\caption{%
		Hyper-parameter values for the experiments in Section~\ref{sec_experiments}. All simulator parameters were	randomized such that they stayed physically plausible.
		We use $n$ as shorthand for $\candsym$ or $\refsym$ depending on the context.
	}
	\label{tab_exp_hparams}
	\rowcolors{2}{gray!25}{white}
\begin{tabular}{ll}
	\rowcolor{gray!50}
	Hyper-parameter                   & Value\\
	\texttt{PolOpt}                   & \acs{PPO}\\
	policy architecture               & \acs{FNN} \num{16}-\num{16} with tan-h\\
	optimizer                         & Adam\\
	learning rate                     & \num{1e-4}\\
	number of iterations $n_{\text{iter}}$   & \num{400}\\
	max. steps per episode $T$        & Ball-Balancer: \num{2000}\\
	                                  & Cart-Pole: \num{2500}\\
	step size $\Delta t$              & \SI{0.002}{\second}\\
	temporal discount $\gamma$        & \num{0.999}\\
	$\lambda$ (andvantage estimation) & \num{0.95}\\
	initial $\candsym$                & \num{5}\\
	initial $\refsym$                 & \num{1}\\
	\texttt{NonDecrSeq}               & $n_{k+1} \gets \floor{n_0(k+1)}$ \\
	rollouts per domain parameter $n_\tau$ & \num{10}\\
	batch size                             & $\ceil*{T/\Delta t} n_\tau n$ steps\\
	number of reference solutions $n_G$    & \num{20}\\
	rollouts per initial state $n_J$       & Ball-Balancer: \num{120}\\
	                                       & Cart-Pole: \num{50}\\
	confidence parameter $\alpha$          & \num{0.05}\\
	threshold of trust $\beta$             & Ball-Balancer: \num{50} \\
	                       & Cart-Pole: \num{60} \\
	number of bootstrap replications $B$   & \num{1000}\\
	\acs{CVaR} parameter $\epsilon$ (\acs{EPOpt})   & \num{0.2}\\
	$r_{min}$     & $10^{-4}$ \\
	$\fQ_{\ballbalancersym}$                  & $\textrm{diag} (1, 1, \num{5e3}, \num{5e3}, \dots$\\
	& $\num{1e-2}, \num{1e-2}, \num{5e-2}, \num{5e-2})$ \\
	$\fR_{\ballbalancersym}$                  & $\mathrm{diag}(\num{1e-3}, \num{1e-3})$\\
% QCP swing-up
%	$c_1, c_2, c_3, c_4$       & \num{1.}, \num{0.1}, \num{0.4}, \num{10.}\\ 
% QCP balancing
	$\fQ_{\cartpolesym}$                  & $\textrm{diag}(\num{10}, \num{1e3}, \num{5e-2}, \num{5e-3})$\\
	$R_{\cartpolesym}$                  & \num{1e-4}\\
% QQ
%	$\fQ_{Q}$                   & $\textrm{diag} (0.1, 1, \num{2e-2}, \num{5e-3})$\\
%	$R_{Q}$                   & \num{3e-3}\\
%	obs. noise mean (for all states) & $0$ \\
	obs. noise std for linear pos & $\SI{5e-3}{\meter}$ \\
	obs. noise std for linear vel & $\SI{0.05}{\meter\per\second}$ \\
	obs. noise std for angular pos & $\SI{0.5}{\degree}$ \\
	obs. noise std for angular vel & $\SI{2.0}{\degree \per \second}$ \\
\end{tabular}
\end{table}

\begin{table*}[thb]
	\centering
	\caption{The domain parameter distributions and derived parameters for the Ball-Balancer (Figure~\ref{fig_Quanser_platforms}~left).
		All parameters were randomized such that they stayed physically plausible. Normal distributions are parameterized with mean and standard deviation, uniform distributions with lower and upper bound.
		The lines separate the randomized domain parameters from the ones depending on these.}
	\label{tab_dom_param_QBB}
	\rowcolors{2}{gray!25}{white}
\begin{tabular}{lll}
	\rowcolor{gray!50}
	\textbf{Parameter}        & \textbf{Distribution}                                             & \textbf{Unit}\\
	gravity constant          & $\distrnormal{g}{\num{9.81},\num{1.962}}$                         & $[\si{kg}]$\\
	ball mass                 & $\distrnormal{m_b}{\num{5e-3},\num{6e-4}}$                        & $[\si{kg}]$\\
	ball radius               & $\distrnormal{r_b}{\num{1.96e-2},\num{3.93e-3}}$                  & $[\si{m}]$\\
	plate length              & $\distrnormal{l_p}{\num{0.275},\num{5.5e-2}}$                     & $[\si{m}]$\\
	kinematic leverage arm    & $\distrnormal{r_{kin}}{\num{2.54e-2},\num{3.08e-3}}$       & $[\si{m}]$\\
	gear ratio                & $\distrnormal{K_g}{70,14}$                                        & $[-]$\\
	gearbox efficiency        & $\distruniform{\eta_g}{\num{1.0},\num{0.6}}$                      & $[-]$\\
	motor efficiency          & $\distruniform{\eta_m}{\num{0.89},\num{0.49}}$                    & $[-]$\\
	load moment of inertia    & $\distrnormal{J_l}{\num{5.28e-5},\num{1.06e-5}}$                  & $[\si{\kilogram\meter\squared}]$\\
	motor moment of inertia   & $\distrnormal{J_m}{\num{4.61e-7},\num{9.22e-8}}$                  & $[\si{\kilogram\meter\squared}]$\\
	motor torque constant     & $\distrnormal{k_m}{\num{7.7e-3},\num{1.52e-3}}$                   & $[\si{\newton\meter\per\ampere}]$\\
	motor armature resistance & $\distrnormal{R_m}{\num{2.6},\num{0.52}}$                         & $[\si{\ohm}]$\\
	motor viscous damping coeff. \wrt load & $\distruniform{B_{\text{eq}}}{\num{0.15},\num{3.75e-3}}$            & $[\si{\newton\meter\second}]$\\
	ball/plate viscous friction coeff.     & $\distruniform{c_v}{\num{5.0e-2},\num{1.25e-3}}$                    & $[-]$\\
	positive voltage threshold x servo     & $\distruniform{V_{thold,x+}}{\num{0.353},\num{8.84e-2}}$     & $[\si{\volt}]$\\
	negative voltage threshold x servo     & $\distruniform{V_{thold,x-}}{\num{-8.90e-2},\num{-2.22e-3}}$ & $[\si{\volt}]$\\
	positive voltage threshold y servo     & $\distruniform{V_{thold,y+}}{\num{0.290},\num{7.25e-2}}$     & $[\si{\volt}]$\\
	negative voltage threshold y servo     & $\distruniform{V_{thold,y-}}{\num{-7.30e-2},\num{-1.83e-2}}$ & $[\si{\volt}]$\\
	offset x servo                         & $\distruniform{\Delta \theta_x}{-5,5}$                              & $[\si{\deg}]$\\
	offset y servo                         & $\distruniform{\Delta \theta_y}{-5,5}$                              & $[\si{\deg}]$\\
	action delay                           & $\distruniform{\Delta t_a}{0,30}$                                   & $[\mathrm{steps}]$\\
	\hline\hline
	kinematic constant                     & $c_{kin} = 2r_{\text{kin}} / l_p$                & $[-]$\\
	combined motor constant                & $A_m = \eta_g K_g \eta_m k_m / R_m$                     & $[\si{\newton\meter\per\volt}]$\\
	combined rotary damping coefficient    & $B_v = \eta_g K_g^2  \eta_m k_m^2 / R_m+ B_{\text{eq}}$ & $[\si{\newton\meter\second}]$\\
	combined rotor inertia                 & $J_{eq} = \eta_g K_g^2 J_m + J_l$                       & $[\si{\kilogram\meter\squared}]$\\
	ball inertia about \acs{CoM}           & $J_b = 2/5 m_b r_b^2$                                   & $[\si{\kilogram\meter\squared}]$\\
	combined ball inertia                  & $\zeta = m_b r_b^2 + J_b$                               & $[\si{\kilogram\meter\squared}]$\\
\end{tabular}

\end{table*}

\begin{table*}[thb]
	\centering
	\caption{The domain parameter distributions and derived parameters for the Cart-Pole (Figure~\ref{fig_Quanser_platforms}~right).
		All parameters were randomized such that they stayed physically plausible. Normal distributions are parameterized with mean and standard deviation, uniform distributions with lower and upper bound.
		The lines separate the randomized domain parameters from the ones depending on these.}
	\label{tab_dom_param_QCP}
	\rowcolors{2}{gray!25}{white}
\begin{tabular}{lll}
	\rowcolor{gray!50}
	\textbf{Parameter}        & \textbf{Distribution}                                & \textbf{Unit}\\
	gravity constant          & $\distrnormal{g}{\num{9.81},\num{1.962}}$            & $[\si{kg}]$\\
	cart mass                 & $\distrnormal{m_c}{\num{0.38},\num{0.076}}$          & $[\si{kg}]$\\
	pole mass                 & $\distrnormal{m_p}{\num{0.127},\num{2.54e-2}}$       & $[\si{kg}]$\\
	half pole length          & $\distrnormal{l_p}{\num{0.089},\num{1.78e-2}}$       & $[\si{m}]$\\
	rail length               & $\distrnormal{l_r}{\num{0.814},\num{0.163}}$         & $[\si{m}]$\\
	motor pinion radius       & $\distrnormal{r_{mp}}{\num{6.35e-3},\num{1.27e-3}}$  & $[\si{m}]$\\
	gear ratio                & $\distrnormal{K_g}{3.71,0}$                          & $[-]$\\
	gearbox efficiency        & $\distruniform{\eta_g}{\num{1.0},\num{0.8}}$         & $[-]$\\
	motor efficiency          & $\distruniform{\eta_m}{\num{1.0},\num{0.8}}$         & $[-]$\\
	motor moment of inertia   & $\distrnormal{J_m}{\num{3.9e-7},\num{0}}$            & $[\si{\kilogram\meter\squared}]$\\
	motor torque constant     & $\distrnormal{k_m}{\num{7.67e-3},\num{1.52e-3}}$     & $[\si{\newton\meter\per\ampere}]$\\
	motor armature resistance & $\distrnormal{R_m}{\num{2.6},\num{0.52}}$            & $[\si{\ohm}]$\\
	motor viscous damping coeff. \wrt load & $\distruniform{B_{eq}}{\num{5.4},\num{0}}$  & $[\si{\newton\second\per\meter}]$\\
	pole viscous friction coeff.           & $\distruniform{B_p}{\num{2.4e-3},\num{0}}$         & $$[\si{\newton\second}]$$\\
%	cart position offset                   & $\distruniform{\Delta \theta_x}{0,5}$              & $[\si{\meter}]$\\
	action delay                           & $\distruniform{\Delta t_a}{0,10}$                  & $[\mathrm{steps}]$\\
	\hline\hline
	pole rotary inertia about pivot point  & $J_p = 1/3 m_p l_p^2$                               & $[\si{\kilogram\meter\squared}]$\\
	combined linear inertia                & $J_{eq} = m_c + (\eta_g K_g^2 J_m) / r_{mp}^2$ & $[\si{\kilogram}]$\\
\end{tabular}

\end{table*}

\end{document}